\documentclass[sigconf]{acmart}

\usepackage{booktabs} 
\usepackage{graphicx}
\usepackage{amsmath}
\usepackage{booktabs}
\usepackage{multirow}
\usepackage{colortbl}
\usepackage{makecell}
\usepackage{diagbox}

\copyrightyear{2023}
\acmYear{2023}
\setcopyright{acmlicensed}
\acmConference[SA Conference Papers'23]{SIGGRAPH Asia 2023 Conference
Papers}{December 12--15, 2023}{Sydney, NSW, Australia}
\acmBooktitle{SIGGRAPH Asia 2023 Conference Papers (SA Conference
Papers'23), December 12--15, 2023, Sydney, NSW, Australia}
\acmPrice{15.00}
\acmDOI{10.1145/3610548.3618142}
\acmISBN{978-8-4007-0315-7/23/12}

\usepackage{xcolor}
\definecolor{myred}{rgb}{0.8,0,0}
\definecolor{mygreen}{rgb}{0,0.8,0}
\usepackage{pifont}

\usepackage[ruled]{algorithm2e} 

\SetAlFnt{\small}
\SetAlCapFnt{\small}
\SetAlCapNameFnt{\small}
\SetAlCapHSkip{0pt}

\begin{document}
\newcommand\methodname{Im4D }

\newcommand\geofunc[0]{\mathbf{F}_{\sigma}}
\newcommand\appfunc[0]{\mathbf{F}_{\mathbf{c}}}
\newcommand\mlpGeo[0]{\mathbf{m}_{\sigma}}
\newcommand\mlpApp[0]{\mathbf{m}_{\mathbf{c}}}

\newcommand\imgfeat[0]{\mathbf{M}}
\newcommand\img[0]{\mathbf{I}}
\newcommand\proj[0]{\mathbf{u}}
\newcommand\projfeat[0]{\mathbf{f}}
\newcommand\difffunc[0]{\textbf{diff}}

\newcommand\volume[0]{\mathbf{V}}
\newcommand\plane[0]{\mathbf{P}}
\newcommand\volumefeat[0]{\mathbf{v}}
\newcommand\planefeat[0]{\mathbf{p}}
\newcommand\colorpixel[0]{\mathbf{C}}

\newcommand\viewdir[0]{\mathbf{d}}
\newcommand\numView[0]{N_v}
\newcommand\numPixel[0]{N_p}
\newcommand\numJointTraining[0]{N_j}
\newcommand\numFinetuneTraining[0]{N_f}
\newcommand\planesizeh[0]{h}
\newcommand\planesizew[0]{w}

\newcommand{\tofig}[1]{\textcolor{cyan}{[TOFIG: #1]}}
\newcommand{\toexp}[1]{\textcolor{orange}{[TOEXP: #1]}}
\newcommand{\modify}[1]{\textcolor{orange}{#1}}

\newcommand{\sida}[1]{\textcolor{cyan}{[Sida: #1]}}

\newenvironment{packed_enum}{
\begin{enumerate}
  \setlength{\itemsep}{1pt}
  \setlength{\parskip}{2pt}
  \setlength{\parsep}{0pt}
}{\end{enumerate}}

\newenvironment{packed_item}{
\begin{itemize}
  \setlength{\itemsep}{1pt}
  \setlength{\parskip}{2pt}
  \setlength{\parsep}{0pt}
}{\end{itemize}}

\newcommand\flickrcc[1]{\small{Photos by Flickr users #1 under \href{https://creativecommons.org/licenses/by/2.0/}{CC BY}}}

\newcommand{\beginsupplement}{%
        \setcounter{table}{0}
        \renewcommand{\thetable}{S\arabic{table}}%
        \setcounter{figure}{0}
        \renewcommand{\thefigure}{S\arabic{figure}}%
     }

\definecolor{colorfirst}{rgb}{.866,.945, 0.831} 
\definecolor{colorsecond}{rgb}{1, 0.98, 0.83} 
\definecolor{colorthird}{rgb}{0.76, 0.87, 0.92} 

\newcommand{\cellfirst}{\cellcolor{colorfirst}}
\newcommand{\cellsecond}{\cellcolor{colorsecond}}
\newcommand{\cellthird}{\cellcolor{colorthird}}

\newcommand{\textfirst}{\colorbox{colorfirst}}
\newcommand{\secondtext}{\colorbox{colorsecond}}
\newcommand{\thirdtext}{\colorbox{colorthird}}

\title{Im4D: High-Fidelity and Real-Time Novel View Synthesis for Dynamic Scenes} 

\author{Haotong Lin}
\author{Sida Peng}
\authornote{Corresponding author.}
\email{haotongl@zju.edu.cn}
\email{pengsida@zju.edu.cn}
\affiliation{%
 \institution{State Key Laboratory of CAD\&CG, Zhejiang University}
 \country{China}
}

\author{Zhen Xu}
\author{Tao Xie}
\author{Xingyi He}
\email{zhenx@zju.edu.cn}
\email{taotaoxie@zju.edu.cn}
\email{xingyihe@zju.edu.cn}
\affiliation{%
 \institution{Zhejiang University}
 \country{China}
 }

\author{Hujun Bao}
\author{Xiaowei Zhou}
\email{bao@cad.zju.edu.cn}
\email{xwzhou@zju.edu.cn}
\affiliation{%
 \institution{State Key Laboratory of CAD\&CG, Zhejiang University}
 \country{China}
 }

\begin{abstract}
This paper aims to tackle the challenge of dynamic view synthesis from multi-view videos. The key observation is that while previous grid-based methods offer consistent rendering, they fall short in capturing appearance details of a complex dynamic scene, a domain where multi-view image-based rendering methods demonstrate the opposite properties. To combine the best of two worlds, we introduce Im4D, a hybrid scene representation that consists of a grid-based \textit{geometry} representation and a multi-view image-based \textit{appearance} representation. Specifically, the dynamic geometry is encoded as a 4D density function composed of spatiotemporal feature planes and a small MLP network, which globally models the scene structure and facilitates the rendering consistency. We represent the scene appearance by the original multi-view videos and a network that learns to predict the color of a 3D point from image features, instead of memorizing detailed appearance totally with networks, thereby naturally making the learning of networks easier. Our method is evaluated on five dynamic view synthesis datasets including DyNeRF, ZJU-MoCap, NHR, DNA-Rendering and ENeRF-Outdoor datasets. The results show that Im4D exhibits state-of-the-art performance in rendering quality and can be trained efficiently, while realizing real-time rendering with a speed of 79.8 FPS for 512x512 images, on a single RTX 3090 GPU. The code is available at \url{https://zju3dv.github.io/im4d}. 
\end{abstract}


\begin{teaserfigure}
    \centering
    \vspace{-2mm}
    \includegraphics[width=0.95\linewidth]{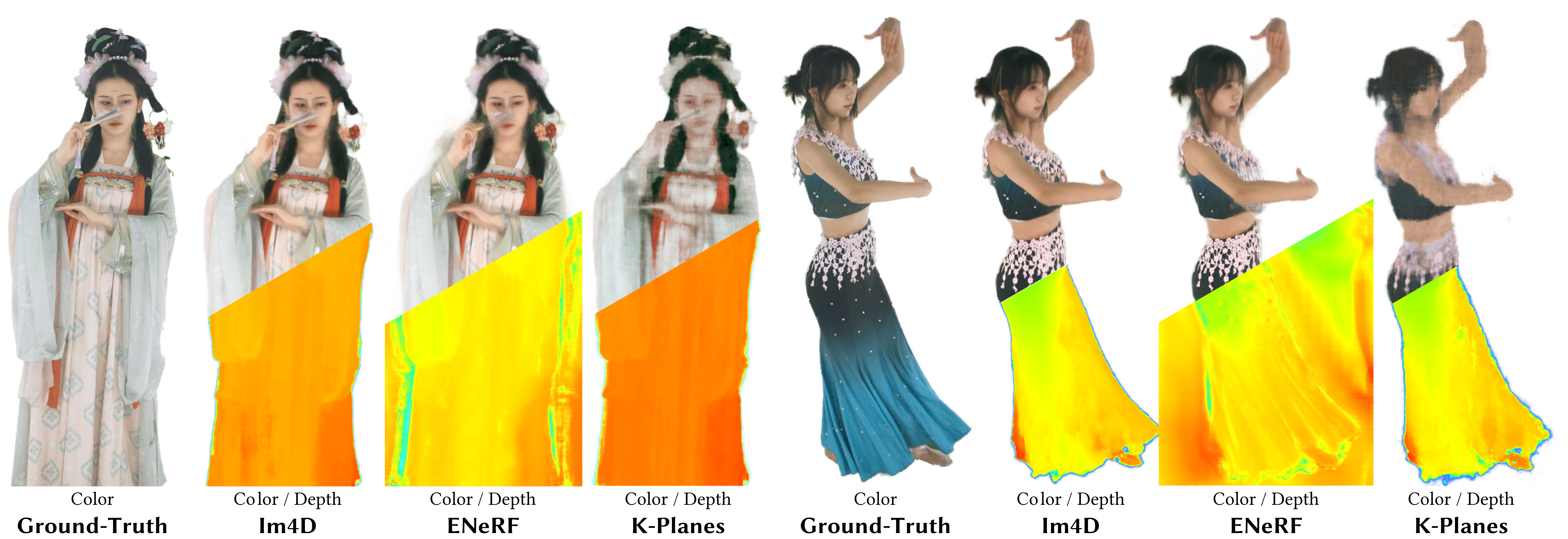}
    \vspace{-2mm}
    \caption{\textbf{Rendering results on the DNA-Rendering \cite{renbody} dataset.} All the rendered depth maps are normalized using the same method with same hyper parameters and then combined with the rendered alpha maps.
    Im4D (our method) produces high-fidelity rendering with high-quality depth results on a long dynamic scene with complex motions. 
    ENeRF \cite{lin2022enerf} struggles to recover correct depth, leading to flicker and ghosting artifacts. K-Planes \cite{fridovich2023k} cannot recover the appearance details on such a difficult scene. Please refer to our video for better visualization.
    }
    \label{fig:teaser}
\end{teaserfigure}

\maketitle

\section{Introduction}
Dynamic view synthesis aims to render novel views of a real-world dynamic scene given input videos, which is a long-standing research problem in computer vision and graphics. It has a variety of applications in film and game production, immersive telepresence, sports broadcasting, etc. 
The key challenge of this problem is to efficiently reconstruct a 4D representation of the dynamic scene from multi-view videos, which allows high-fidelity (i.e., photo-realistic and multi-view consistent) and real-time rendering at arbitrary viewpoints and time.

Recent methods \cite{mildenhall2020nerf} have achieved great success in novel view synthesis for static scenes using implicit scene representations, which brings new insights to the field of dynamic view synthesis. 
Some methods (e.g., \cite{li2021neural, xian2021space}) have been proposed to enhance NeRF's \cite{mildenhall2020nerf} MLP by incorporating time as an additional input, enabling the representation of radiance fields in dynamic scenes.
By employing the volume rendering technique with a series of training strategies, DyNeRF \cite{li2021neural} is able to achieve realistic rendering after training for one week on 8xV100 GPUs. 
To address the challenge of training efficiency, recent methods \cite{fridovich2023k,fang2022fast} draw inspiration from \cite{muller2022instant,chen2022tensorf} and incorporate explicit optimizable embeddings into the implicit representation, significantly accelerating the training speed.
However, as depicted in Fig. \ref{fig:teaser}, these methods encounter difficulties in capturing the appearance details of complex dynamic scenes. 
Some possible solutions are to increase the model size or divide the sequence and process it using multiple models, which will lead to a linear increase in training time and model size.

We observe that 2D videos faithfully record the appearance of scenes and video compression techniques have been well studied and standardized \cite{sullivan2012overview}, allowing very efficient storage and transmission.
Motivated by this, we propose Im4D, a novel hybrid scene representation for dynamic scenes, which consists of a grid-based 4D geometry representation and a multi-view image-based appearance representation, for efficient training and high-fidelity rendering of complex dynamic scenes.
Specifically, given a spatial point at a specific time step, we fetch the corresponding feature from explicit spatiotemporal feature planes \cite{fridovich2023k} and then regress the density from this feature with a small MLP.
For the appearance part, we first feed the nearby views of the rendered view at the time step into a CNN network to obtain feature maps. Then, we project the spatial point onto these feature maps to obtain pixel-aligned features. Finally, we utilize a small MLP to predict the color from these features.
This representation can be rendered using the volume rendering technique.
Unlike previous methods \cite{fridovich2023k,fang2022fast} that \textit{memorize} the radiance of each space-time point along each direction (in $\mathbb{R}^6$ space), our method learns \textit{inferring} the radiance from input image features.
We experimentally show that the proposed method achieves faster training and better rendering quality, indicating that our strategy effectively reduces the learning burden of the network.

In addition to boosting the training speed and rendering quality, Im4D inherently ensures better cross-view rendering consistency.
Previous methods \cite{wang2021ibrnet,lin2022enerf} also utilize an image-based rendering representation, but they simultaneously predict the geometry and appearance for each rendering. Their nature of \textit{per-view} reconstruction instead of \textit{global} reconstruction cannot ensure rendering consistency between viewpoints.
With the global geometry representation, Im4D achieves better cross-view rendering consistency than these methods as shown in our video.

We evaluate our method on several commonly used benchmarks for dynamic view synthesis, including the ZJU-MoCap \cite{peng2021neural}, NHR \cite{wu2020multi}, DyNeRF \cite{li2021neural} and DNA-Rendering \cite{renbody} datasets. Our method consistently demonstrates state-of-the-art performance across all of these datasets, achieved with a training time of a few hours.
Furthermore, we demonstrate that our representation can be rendered in real-time, capable of rendering 512x512 images at a speed of 79.8 FPS on a single RTX 3090 GPU. 
We further validate the versatility of our method on the ENeRF-Outdoor \cite{lin2022enerf} dataset. This dataset is particularly challenging for dynamic view synthesis, as it comprises large motions within complex outdoor scenes.

In summary, this work makes the following contributions:
1) We propose Im4D, a novel hybrid representation for dynamic scenes, which consists of a grid-based 4D geometry representation and a multi-view image-based appearance representation.
2) We conduct extensive comparisons and ablations on several datasets, demonstrating that our method achieves state-of-the-art performance in terms of rendering quality with training in a few hours.
3) We demonstrate that the proposed representation can be rendered in real-time on the ZJU-MoCap dataset.

\begin{figure*}
    \centering
    \includegraphics[width=0.95\linewidth]{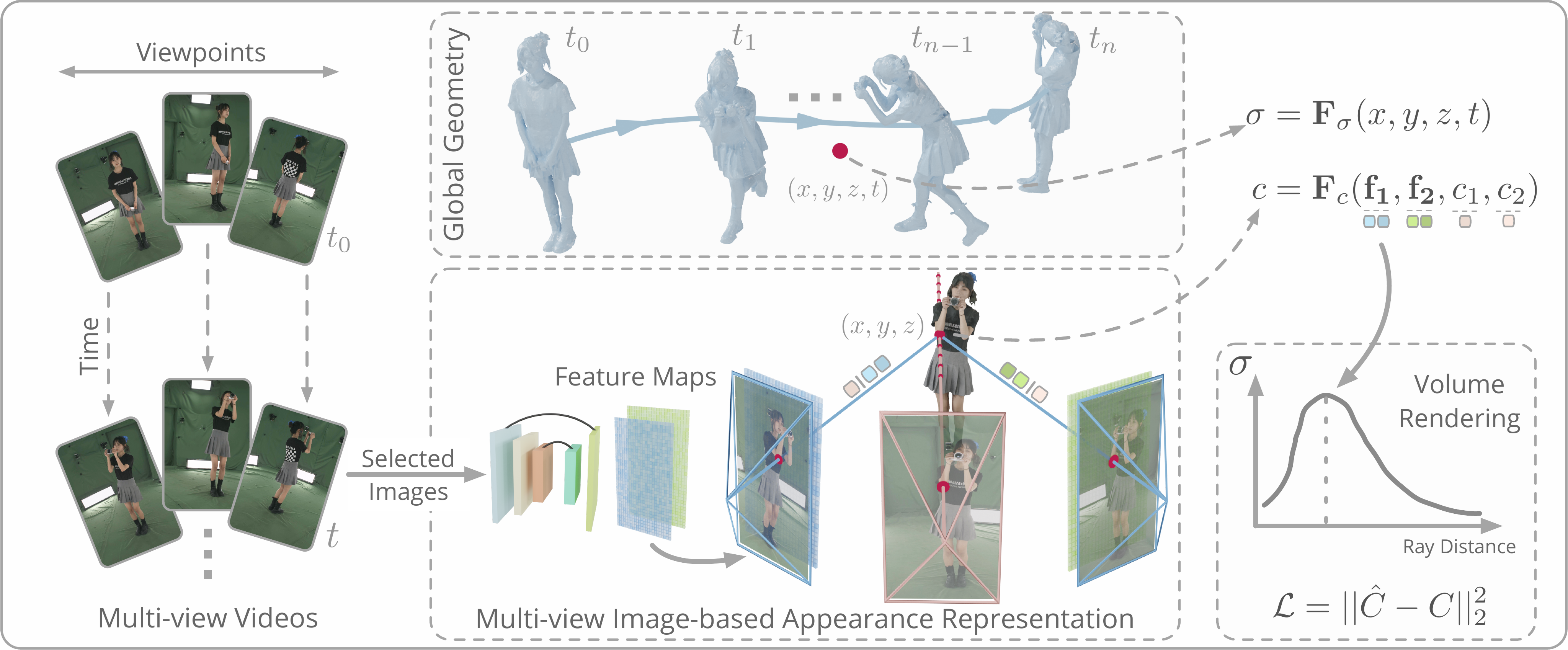}
    \vspace{-1mm}
    \caption{Overview of Im4D.
Given a set of multi-view videos, the proposed method aims to reconstruct a 3D model capable of rendering photorealistic images at arbitrary viewpoints and time steps.  
The proposed method models the geometry with a \textit{global} 4D density function. This function consists of a small MLP and a 4D space structure storing optimizable features.
The appearance part is represented with a multi-view image-based appearance model, which learns to predict the color of a 3D point from image features extracted from selected images (the input images closest to the rendering view).
}
    \label{fig:main_fig}
    \vspace{-2mm}
\end{figure*}

\section{Related Works}
\paragraph{Novel View Synthesis of Static Scenes.} 
Novel view synthesis has traditionally been approached via several paradigms, including light field-based methods \cite{gortler1996lumigraph, levoy1996light, davis2012unstructured}, multi-view images \cite{zitnick2004high, chaurasia2013depth,flynn2016deepstereo, kalantari2016learning, penner2017soft,chris2001unstructured}, and multi-plane images \cite{li2020crowdsampling,tucker2020single,mildenhall2019local,zhou2018stereo,srinivasan2019pushing,szeliski1998stereo}. Recently, a new paradigm has emerged in the form of neural representations \cite{hedman2018deep,sitzmann2019deepvoxels, liu2019neural, lombardi2019neural, shih20203d,liu2019learning,jiang2020sdfdiff, wizadwongsa2021nex,suhail2022cvpr,sitzmann2021light,attal2022learning, kellnhofer2021neural} for novel view synthesis. NeRF \cite{mildenhall2020nerf} represents scenes as neural radiance fields using a Multi-Layer Perceptron (MLP), yielding impressive rendering results.
NeRF requires a lengthy per-scene optimization.
To avoid the training burden, several methods \cite{yu2020pixelnerf,wang2021ibrnet,liu2021neural,chen2021mvsnerf,johari2021geonerf,chibane2021stereo} propose to use an MLP to decode radiance fields from pixel-aligned image features from nearby images. These methods can be quickly fine-tuned to new scenes by pre-training a CNN to extract feature maps over large datasets. 
There exist some works \cite{sun2020light,kopanas2021point} that study how to select nearby images and develop a smooth fading strategy to avoid temporal instability.
Some methods \cite{muller2022instant,chen2022tensorf,yu2021plenoxels,sun2021direct} design hybrid or explicit structures to store optimizable features or latent representations (e.g., spherical harmonics), achieving rapid reconstruction.
To improve the rendering speed, some methods \cite{liu2020neural,yu2021plenoctrees,garbin2021fastnerf,hedman2021baking,reiser2021kilonerf} represent or cache the radiance fields into efficient structures.
Another category of methods \cite{neff2021donerf} has also achieved significant acceleration by using depth to expedite rendering.
More recently, several methods \cite{chen2022mobilenerf,wan2023learning} design representations based on polygonal mesh and leverage graphics pipelines to achieve real-time rendering effects.
A few studies \cite{barron2021mip,barron2022mip,barron2023zip,turki2022mega,tancik2022block} have also analyzed issues such as anti-aliasing inherent in NeRF representation 
In addition to NeRF-based methods, several methods \cite{Riegler2020FVS,riegler2021stable} have improved rendering quality or reconstruction quality by incorporating multi-view image-based rendering appearance models into surface models.
\cite{bergman2021fast} explores a similar idea that uses multi-view images and an MLP for appearance and geometry, respectively.

\paragraph{Novel View Synthesis of Dynamic Scenes}
Early works \cite{dou2016fusion4d,newcombe2015dynamicfusion,yu2018doublefusion,orts2016holoportation} predominantly used explicit surface models for dynamic scenes. They \cite{collet2015high} rely on depth sensors and multi-view stereo techniques to capture per-view depth before consolidating it into the scene geometry.
In a different approach, NeuralVolumes \cite{lombardi2019neural} use volume rendering to reconstruct 4D scenes from color images. However, its usage of 3D volumes imposed limitations on achieving high-resolution results. In response, some works \cite{peng2023representing,lombardi2021mixture} seek to convert the 3D volume into a 2D representation, as MLP maps and UV maps, respectively. While they provide a lightweight solution that is capable of real-time rendering, they also demand extended training periods.
Progressing further, a collection of studies extended NeRF to accommodate dynamic scenes, paving the way for high-resolution rendering. 
These methods generally incorporated a time variable into the NeRF's MLP \cite{xian2021space,li2021neural,lin2023neural}, scene flows \cite{li2020neural,du2021nerflow,gao2021dynamic,you2023decoupling} or use deformable fields \cite{pumarola2020d,park2021nerfies,park2021hypernerf,zhang2021editable}. Despite the promising results, these techniques necessitated considerable training resources.
To optimize training efficiency, recent methodologies \cite{fridovich2023k,fang2022fast,cao2023hexplane,shao2022tensor4d,wang2022fourier,attal2023hyperreel,gan2022v4d} introduced optimizable embeddings into the implicit representation, resulting in training speed-ups. However, as demonstrated by our experiments, K-Planes \cite{fridovich2023k} still falls short of delivering clear rendering results on challenging dynamic scenes.
Some recent methods \cite{song2023nerfplayer,li2022streaming,wang2023inv} also explore the streaming representation of dynamic scenes.
Another line of works such as \cite{lin2022enerf,wang2021ibrnet} deploy multi-view images to represent 4D scenes, offering high-quality rendering and efficient training. Nonetheless, due to their inherent per-view reconstruction nature, they are unable to guarantee cross-view rendering consistency.
In contrast to these, our hybrid dynamic scene representation ensures rendering consistency by employing a global grid-based geometry representation. At the same time, we use multi-view images for appearance representation, achieving high-quality rendering.
More recently, a concurrent work \cite{icsik2023humanrf} has similar findings that the grid-based representations significantly degrade when dealing with long and complex dynamic scenes. They address this issue by adaptively dividing the dynamic scene into much shorter segments.

\section{Method}
\label{sec:method}

Given multi-view videos of a dynamic scene, our objective is to construct a time-varying 3D model capable of generating photorealistic images from any perspective and any time (discrete time steps of input frames). 
We propose Im4D, a novel scene representation, which combines the strengths of both \textit{global} methods \cite{mildenhall2020nerf,fridovich2023k}, with their cross-view rendering consistency, and multi-view image-based rendering (\textit{per-view}) methods \cite{wang2021ibrnet}, recognized for their high-quality rendering and fast training \cite{lin2022enerf}.
Im4D consists of two parts: a global grid-based dynamic geometry representation (Sec. \ref{sec:method_geo}) and a multi-view image-based appearance representation (Sec. \ref{sec:method_app}).
As shown in Fig. \ref{fig:main_fig}, we represent the density fields with a continuous function that takes $(x, y, z, t)$ as input.
The appearance is modeled using a multi-view image-based rendering model, which infers the color of a 3D point from nearby multi-view images at the given time.
The proposed representation is optimized using RGB images and can be rendered in real-time. (Sec. \ref{sec:method_opt}).

\subsection{Grid-based Dynamic Geometry}
\label{sec:method_geo}

In this section, we seek to represent the dynamic geometry of a scene as a time-varying 3D model to achieve inter-view rendering consistency.
Inspired by recent progress in static 3D reconstruction \cite{mildenhall2020nerf,yu2021plenoxels,muller2022instant}, we represent the dynamic geometry of a scene as a grid-based model.

In static 3D reconstruction, the geometry of a scene is represented as a 3D vector-valued function that takes a 3D position coordinate as input and outputs a volume density.
To consistently represent the dynamic geometry, we extend the static geometry function to a 4D vector-valued function, where the input becomes a 3D coordinate $(x, y, z)$ and time $t$:
\begin{equation}
    \sigma = \geofunc (x, y, z, t).
\label{eq:geofunc}
\end{equation}
In practice, we utilize a hybrid representation to implement $\geofunc$, which interpolates the $(x, y, z, t)$ in the 4D volume $\volume$ to obtain $\volumefeat(x, y, z, t)$ and then a small MLP network $\mlpGeo$ maps the feature $\volumefeat(x, y, z, t)$ to a scalar density $\sigma$.
Inspired by \cite{chan2022efficient,chen2022tensorf}, we decompose the 4D volume $\volume$ into six orthogonal planes $\{\plane_i | i \in {xy, xz, yz, xt, yt, zt} \}$ to maintain efficiency in storage. Thus the feature $\volumefeat(x, y, z, t)$ can be defined as the aggregation of features $\{\planefeat_i = \textbf{interp}(\plane_i, i) | i \in {xy, xz, yz, xt, yt, zt} \}$. For simplicity, we simply use concatenation as the aggregation function. The final dynamic geometry model can be formulated as:
\begin{equation}
    \sigma = \mlpGeo(\planefeat_{xy} \oplus \planefeat_{xz} \oplus \planefeat_{yz} \oplus \planefeat_{xt} \oplus \planefeat_{yt} \oplus \planefeat_{zt}).
\end{equation}

The proposed grid-based dynamic geometry function is global and continuous in both space and time. With the volume rendering technique, the proposed model inherently ensures cross-view rendering consistency.
Rather than representing the geometry as a global function, previous multi-view image-based methods \cite{wang2021ibrnet,lin2022enerf} predict the geometry for a novel viewpoint from the nearby multi-view images. This can be regarded as \textit{per-view} reconstruction instead of the \textit{global} reconstruction, which makes it difficult to achieve consistent rendering, resulting in flickering artifacts.
Some concurrent works \cite{fridovich2023k,shao2022tensor4d,cao2023hexplane} propose similar representations to represent the dynamic scene. 
However, they represent both geometry and appearance as a global function, which struggles to achieve high-fidelity rendering. 
In contrast, we only represent the geometry as a global 4D function. 
Next, we will introduce how to efficiently represent the appearance of a dynamic scene.

\subsection{Multi-view Image-based Appearance}
\label{sec:method_app}
This section delves into finding a representation for high-fidelity appearance model. 
Specifically, we aim to predict the radiance of the space-time point $(x, y, z, t)$ along the view direction $\viewdir$. 
A straightforward approach would be to extend Eq. \ref{eq:geofunc} to predict an additional geometry feature, and then use an MLP that takes the geometry feature and view direction $\viewdir$ as inputs to predict radiance. 
This method is commonplace in static scenes \cite{mildenhall2020nerf,muller2022instant}. However, the same strategy, when applied to complex dynamic scenes where the information becomes more abundant and complex, tends to render blurry results as shown in Fig. \ref{fig:teaser}.
To address this issue, we propose a multi-view image-based appearance model. 
Rather than modeling the appearance as an MLP with 4D explicit structure, we predict the color from the image features, which are projected from multi-view image feature maps of the 4D space-time point.
In this manner, the appearance model only needs to learn the problem of \textit{inferring}, rather than \textit{memorizing} the radiance values at each point along each viewpoint in space and time (in $\mathbb{R}^6$ space). 

Specifically, to render an image from a novel space-time viewpoint, we first select $\numView$ input images from the input multiple videos that are spatially close to the desired viewpoint at the given time. 
The spatial proximity is defined as the distance between camera positions.
Then we use a 2D UNet \cite{ronneberger2015u} to extract features from the $\numView$ selected images, resulting in $\{\imgfeat_i  | i \in [0, \numView) \}$, 
where $\imgfeat_i \in \mathbb{R}^{C_i \times H_i \times W_i}$ is the feature map extracted from the $i$-th view, and $C_i$, $H_i$, $W_i$ denote the number of channels, height, and width of the feature map, respectively. 
Next, we project the 3D point $(x, y, z)$ to the $\numView$ views to obtain the corresponding 2D coordinates $\{\proj_i | i \in [0, \numView) \}$, where $\proj_i \in \mathbb{R}^{2}$ is the projected 2D coordinates of the $i$-th view. 
Then we use the bilinear sampling to sample the feature map $\imgfeat_i$ at the projected coordinates $\proj_i$ to obtain the projected features $\{\projfeat_i | i \in [0, \numView) \}$ and pixel colors $\{ \mathbf{c}_{i} | i \in [0, \numView]\}$. The projected features $\{\projfeat_i | i \in [0, \numView) \}$ are aggregated to obtain the fused feature $\projfeat$. For simplicity, we use variance and mean as the aggregation function, which is formulated as:
\begin{equation}
    \projfeat = \textbf{VAR}( \{\projfeat_i | i \in [0, \numView) \}) \oplus \textbf{MEAN}( \{\projfeat_i | i \in [0, \numView) \}).
\end{equation}
The color $\mathbf{c}$ of point $(x, y, z, t)$ is predicted using a pointnet-like \cite{qi2017pointnet} structure similar to ENeRF~\cite{lin2022enerf}, which is invariant to the order of the input features:
\begin{equation}
    \mathbf{c} = \frac{\sum_{i}^{\numView} \mlpApp(\projfeat, \projfeat_i, \difffunc(\viewdir, \viewdir_i)) * \mathbf{c}_i}{\sum_{i}^{\numView} \mlpApp(\projfeat, \projfeat_i, \difffunc(\viewdir, \viewdir_i)) },
\end{equation}
where $\viewdir_i$ is the view direction from $i$-th view camera position to point $(x, y, z)$ and $\difffunc(\viewdir, \viewdir_i)$ is the normalized angle difference between $\viewdir$ and $\viewdir_i$.


\begin{figure*}
    \centering
    \includegraphics[width=0.95\linewidth]{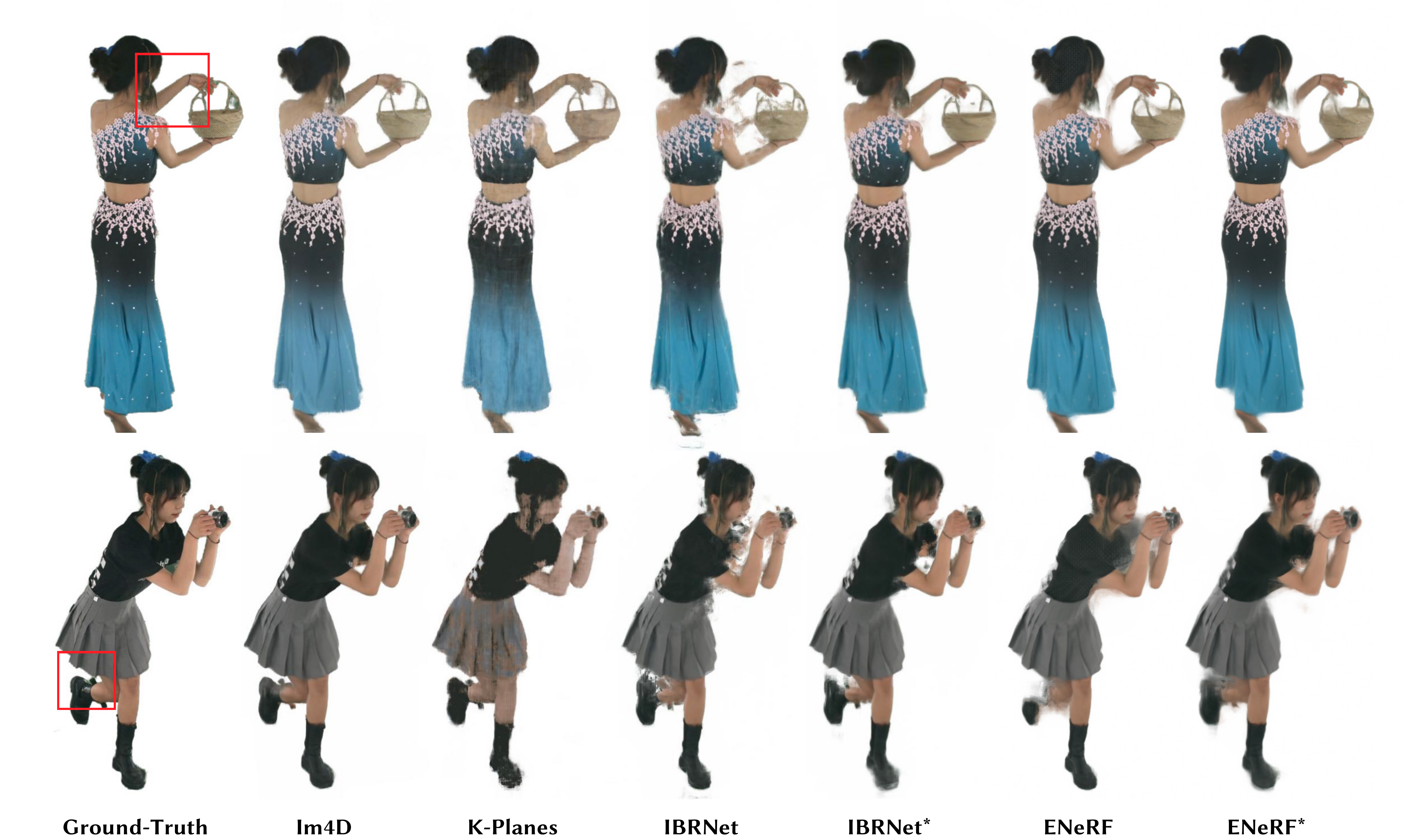}
    \vspace{-2mm}
    \caption{\textbf{Qualitative comparison of image synthesis results on the DNA-Rendering dataset.} The upperscript * implies that the results are obtained with extensive per-scene fine-tuning. IBRNet and ENeRF often produce artifacts in thin structures or occluded regions. Our method produces high-fidelity rendering and superior results in these regions, owing to the global geometry representation. K-Planes struggles to recover the appearance details.
    }
    \label{fig:comp_renbody}
    \vspace{-3mm}
\end{figure*}

\subsection{Efficient Training and Rendering}
\label{sec:method_opt}
\paragraph{Optimization.}
In the previous sections, we described how to obtain color and view-dependent radiance using the proposed scene representation. 
Given the estimated density and radiance fields, we employ the volume rendering technique \cite{mildenhall2020nerf} to render pixel colors. 
We optimize our scene representation using the mean squared error (MSE) loss with color images. 
In addition to the color loss, we follow \cite{yu2021plenoxels,fridovich2023k} and use the total variation (TV) loss to regularize the explicit feature planes $\{\plane_i | i \in \{xy, xz, yz, xt, yt, zt\}\}$. 
We detail our loss formulation in the supplementary material.

\paragraph{Efficient Training.}
In Table. \ref{tab:comp_zju_nhr}, our experiments show that our approach with an image-based appearance model obtains higher rendering quality at 100K iterations than the one without image-based appearance model training with 480k iterations.
To further improve the training speed, we exploit the property of a multi-view image-based appearance model and design a specialized training strategy.
The basic idea is that the appearance model can be quickly trained as it is an inferring model.
We disable the training of the appearance model to avoid the backward time, which accelerates one training iteration.
Specifically, we first jointly train the image feature network with the geometry model for $N_j$ iterations and then finetune the image feature network every $N_f$ iterations.
In practice, we set $N_j$ to 5000 and $N_f$ to 20 in all experiments.

\paragraph{Efficient Rendering}
Previous multi-view image-based rendering methods, such as ENeRF \cite{lin2022enerf}, demonstrate that estimating the depth of novel views as coarse geometry to guide sampling can accelerate rendering.
In our approach, we have a global geometry model, which enables us to accelerate rendering by precomputing the global coarse geometry.
Specifically, we compute a binary field that indicates whether each voxel is occupied or empty. This binary field is then used to guide sampling, allowing us to skip sampling in empty regions and thus speeding up the rendering process.
We use NerfAcc\cite{li2023nerfacc} to implement the described efficient rendering strategy.
The storage requirement for this binary field is remarkably small. For example, storing a binary field for 300 frames of size 64x64x128 in the ZJU-MoCap dataset only requires 18.75MB, and can be losslessly compressed further to 1.1MB.
This acceleration technique can be more efficient than ENeRF as we only need to compute the coarse geometry (the \textit{global} binary field) for once, while ENeRF needs to estimate the coarse geometry (\textit{per-view} depth) for each rendering.
(Ours 79 FPS v.s. ENeRF 51FPS for rendering images of 512x512 on the ZJU-MoCap.)


\begin{table}
    \tabcolsep=0.2cm
    \begin{center}
    \caption{\textbf{Quantitative comparison on the DNA-Rendering dataset.} 
    The zero training time means that methods are tested with their released pre-trained model. 
    We observe that our method typically converges after 140k training iterations and report the corresponding training time. We additionally report our results at the 20k iterations to show the training efficiency. 
    For other methods, we identify a specific training step to ensure convergence, and then report the training time for these methods at this fixed training step. (120k and 480k training iterations for K-Planes).
    The best and second-best results are highlighted \textfirst{green} and \secondtext{yellow}, respectively.}
    \vspace{-2mm}
    \label{tab:comp_renbody}
    \resizebox{0.95\columnwidth}{!}{
    \begin{tabular}{l|c|ccc}
    \Xhline{3\arrayrulewidth}
    & Training time (hour) & PSNR $\uparrow$ & SSIM $\uparrow$ & LPIPS $\downarrow$  \\
    \Xhline{3\arrayrulewidth}
    \multirow{2}{*}{K-Planes} & 2   & 26.34  & 0.943 & 0.134  \\ 
     & 8 & 27.45 & 0.952  & 0.118  \\ \hline
    \multirow{2}{*}{IBRNet} & 0 & 25.31 & 0.954 & 0.106  \\
    & 3.5 & 27.85 & 0.967 & 0.081 \\ \hline
    \multirow{2}{*}{ENeRF} & 0 & 26.85 & 0.966 & 0.073 \\
    & 3.5 & \cellsecond 28.07 & \cellsecond 0.968 & \cellsecond 0.066 \\ \hline
    \multirow{2}{*}{Im4D} & 0.51 & 27.14 & 0.961 & 0.083 \\
     & 3.33 &\cellfirst 28.99 &\cellfirst 0.973 & \cellfirst 0.062 \\
    \Xhline{3\arrayrulewidth}
    \end{tabular}
    }
    \end{center}
    \vspace{-3mm}
\end{table}

\section{Experiments}
\label{sec:exp}

\subsection{Datasets and Metrics}
\label{sec:metric}
We evaluate our method on 4 datasets for dynamic view synthesis, including DNA-Rendering \cite{renbody}, ZJU-MoCap \cite{peng2021neural}, NHR \cite{wu2020multi}, and DyNeRF \cite{li2021neural}. 
DNA-Rendering contains dynamic objects and humans. 
Its videos are recorded at 15 FPS and each clip lasts for 10 seconds. 
DNA-Rendering is very challenging due to the complex clothing textures and movements featured in the videos.
The NHR dataset features a frame rate of 30FPS over a 3.5-second duration, while the ZJU-MoCap dataset runs at 50FPS over 6 seconds.
We evaluate our method on 4 sequences for both NHR and DNA-Rendering datasets with all frames, using 90\% of the views for training and the remaining views for evaluation. 
We conduct experiments on 9 sequences for the ZJU-MoCap dataset.
The DyNeRF dataset contains dynamic foreground objects and complex background scenes. 
It is captured by 15-20 cameras of 30FPS@10 seconds. We conduct quantitative experiments on one sequence, using one view as the test set and the remaining views as training data.
All images are resized to a ratio of 0.5 for training and testing.
We also include the view synthesis results of our method on the ENeRF-Outdoor dataset \cite{lin2022enerf} in the supplementary video.

To evaluate the dynamic view synthesis task, NeuralBody \cite{peng2021neural} suggests evaluating metrics only for the dynamic regions. For MoCap datasets, dynamic regions can be obtained by projecting a predefined 3D bounding box of the person onto the images. We follow this definition for the NHR, ZJU-MoCap, and DNA-Rendering datasets.
For the DyNeRF dataset, previous methods \cite{fridovich2023k,li2021neural} directly evaluated the entire image.
However, we argue that this approach is unreasonable due to the very small motion in this dataset. We introduce a new evaluation approach to specifically evaluate the dynamic regions. 
Specifically, for a 10-second video, we take the first frame of each second as test frames.
For each test frame, we calculate the average frame for that second.
Then, we identify several non-overlapping patches with the highest average pixel differences between the test frame and the average image.
Only these patches are used for evaluation.
We recommend taking 6 patches with a patch size of 128 for the DyNeRF dataset. 

\begin{table}
    \begin{center}
    \caption{\textbf{Quantitative comparison on the DyNeRF dataset.}
We include quantitative results for both entire image and dynamic regions (entire/dynamic). 
The description of this evaluation setting can be found in Sec. \ref{sec:metric}.
    }
    \label{tab:comp_dynerf}
    \resizebox{0.95\columnwidth}{!}{
    \begin{tabular}{l|c|ccc}
    \Xhline{3\arrayrulewidth}
    & Training time & PSNR $\uparrow$ & SSIM $\uparrow$ & LPIPS $\downarrow$  \\
    \Xhline{3\arrayrulewidth}
    \multirow{2}{*}{K-Planes} & 0.8 &  30.78/27.29 & 0.953/0.887  & 0.218/0.379  \\ 
     & 2 &  31.61/29.62 &  0.961/0.916  &  0.182/0.306  \\ \hline
    IBRNet & 3 &  \cellsecond 31.52/31.91 & \cellsecond 0.963/0.956 & \cellfirst 0.169/0.144 \\ \hline
    Im4D & 0.46 & \cellfirst 32.58/32.05 & \cellfirst 0.971/0.956 &  \cellsecond 0.208/0.170 \\
    \Xhline{3\arrayrulewidth}
    \end{tabular}
    }
    \end{center}
\end{table}
 
\begin{figure}
    \centering
    \includegraphics[width=1\columnwidth]{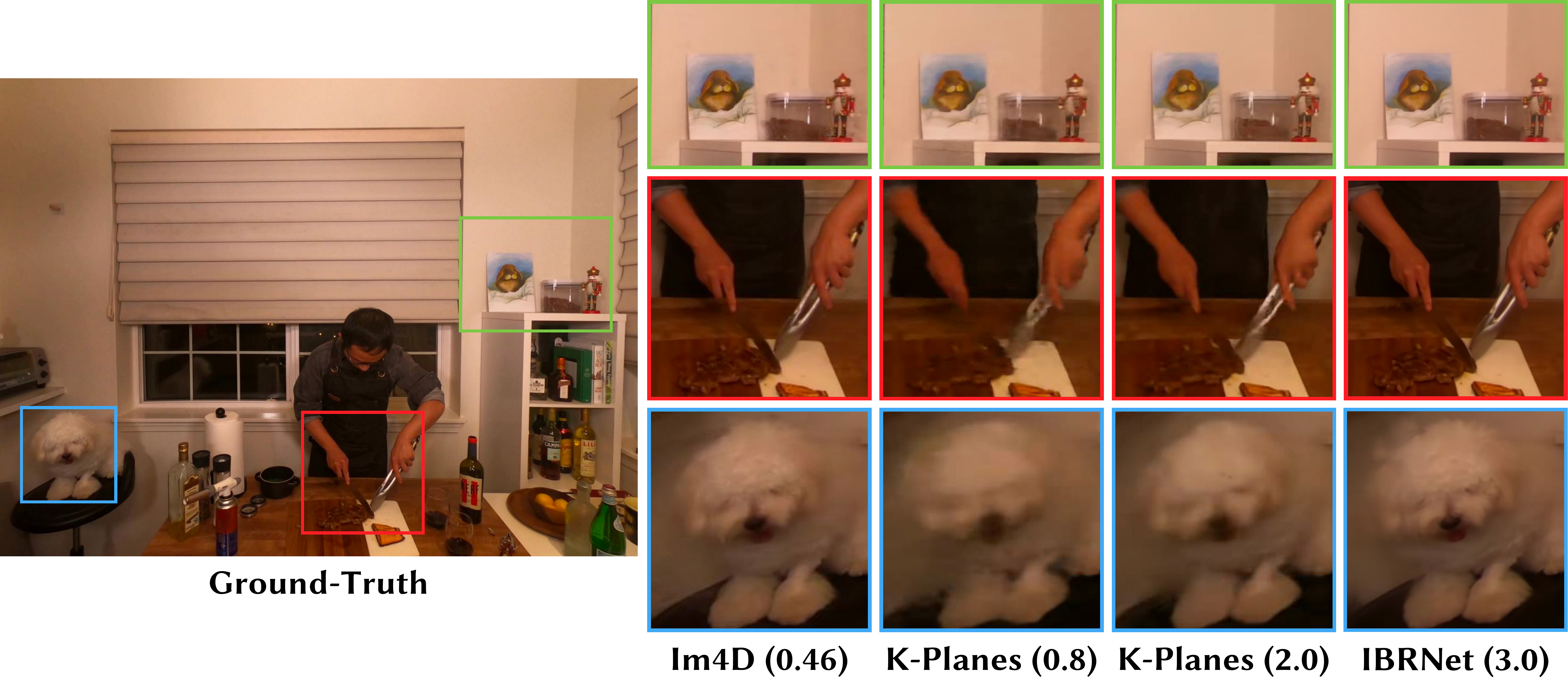}
    \vspace{-3mm}
    \caption{\textbf{Qualitative comparison on the DyNeRF dataset.} 
    }
    \label{fig:comp_dynerf}
    \vspace{-3mm}
\end{figure}

\subsection{Comparisons with State-of-the-art Methods}
\paragraph{Comparison methods.}
We make comparisons with several open-sourced SOTA methods. These methods can be divided into two categories: 1) methods that optimize \textit{per-view} radiance fields (multi-view image-based methods), including IBRNet \cite{wang2021ibrnet} and ENeRF \cite{lin2022enerf}. 2) methods that optimize \textit{global} scene representations, including DyNeRF \cite{li2021neural3d}, K-Planes \cite{fridovich2023k} and MLP-Maps \cite{peng2023representing}. We use the official implementation of them. Our comparison setting strictly follows MLP-Maps and some quantitative results are taken from it. 

\paragraph{Comparison results.}
We first make comparisons with the state-of-the-art methods on the DNA-Rendering dataset. 
The corresponding qualitative and quantitative results are shown in Fig. \ref{fig:comp_renbody} and Table \ref{tab:comp_renbody}, respectively.
As shown in the results, K-Planes struggles to recover the appearance details even after training for eight hours. Other image-based methods (IBRNet, ENeRF and Our method) can recover the high-quality appearance, but IBRNet and ENeRF have artifacts in the occluded regions and regions including thin structures (e.g., legs and hands). We include more comparison results in the supplementary video, where the artifacts for ENeRF and IBRNet can be observed more clearly.
We also make quantitative comparisons with the state-of-the-art methods on the ZJU-MoCap and NHR datasets.
As shown in Table \ref{tab:comp_zju_nhr}, our method outperforms all the other methods in terms of all metrics on both datasets. 
Table \ref{tab:comp_dynerf} and Fig. \ref{fig:comp_dynerf} provide quantitative and qualitative comparison results on the DyNeRF dataset, respectively.
We include the per-scene breakdown in the supplementary material.


\begin{table}
    \tabcolsep=0.1cm
    \begin{center}
    \caption{\textbf{Quantitative comparison on ZJU-MoCap and NHR datasets.} The results of DyNeRF* and MLP-Maps* are taken from MLP-Maps. 
    The zero training time means that the method is tested with their released pre-trained models. 
    Please refer to the supplementary for qualitative results.
    }
    \vspace{-2mm}
    \label{tab:comp_zju_nhr}
    \resizebox{\columnwidth}{!}{
    \begin{tabular}{l|c|ccc|ccc}
    \Xhline{3\arrayrulewidth}
    & \multirow{2}{*}{ \shortstack{Training \\ time (hour) }} &  \multicolumn{3}{c|}{ZJU-MoCap} & \multicolumn{3}{c}{NHR} \\
    &   & PSNR & SSIM & LPIPS & PSNR & SSIM & LPIPS \\
    \hline
    DyNeRF* & >24 & 29.88 & 0.959 & 0.087 & 30.87 & 0.943 & 0.118 \\  \hline
    MLP-Maps* & >24 & \cellsecond 30.17 & \cellsecond 0.963 & 0.068 & 32.20 & 0.953 & 0.080 \\ \hline
    \multirow{2}{*}{K-Planes} & 2 & 29.50 & 0.956  & 0.107 & 30.41 & 0.943 & 0.137 \\
                              & 8 & 30.16 & 0.962  & 0.082 & 32.93 & 0.958 & 0.101 \\ \hline
    \multirow{2}{*}{IBRNet} & 0 & 27.94 & 0.935  & 0.126 & 28.63 & 0.935 & 0.113\\ 
     & 2.5 & 29.40 & 0.956  & 0.084 &\cellsecond 33.53 & \cellsecond 0.965 & 0.077\\ \hline
    \multirow{2}{*}{ENeRF} & 0 & 29.10  & 0.959 & 0.051  & 26.39 & 0.931 & 0.088  \\
     & 2.5 & 29.21 & 0.959  & \cellsecond 0.049 & 30.56  & 0.954  & \cellsecond 0.074 \\ \hline
    \multirow{2}{*}{Im4D} & 0.49     &  30.07 & 0.964 & 0.061 & 32.40 & 0.962 & 0.074 \\
                          & 2.29 & \cellfirst 30.49 & \cellfirst 0.966 &\cellfirst 0.049 & \cellfirst 33.72 & \cellfirst 0.970 & \cellfirst 0.055  \\
    \Xhline{3\arrayrulewidth}
    \end{tabular}
    }
    \end{center}
    \vspace{-3mm}
\end{table}
  
\subsection{Ablations and Analysis}
\label{sec:ablation}
\begin{figure}
    \centering
    \includegraphics[width=\columnwidth]{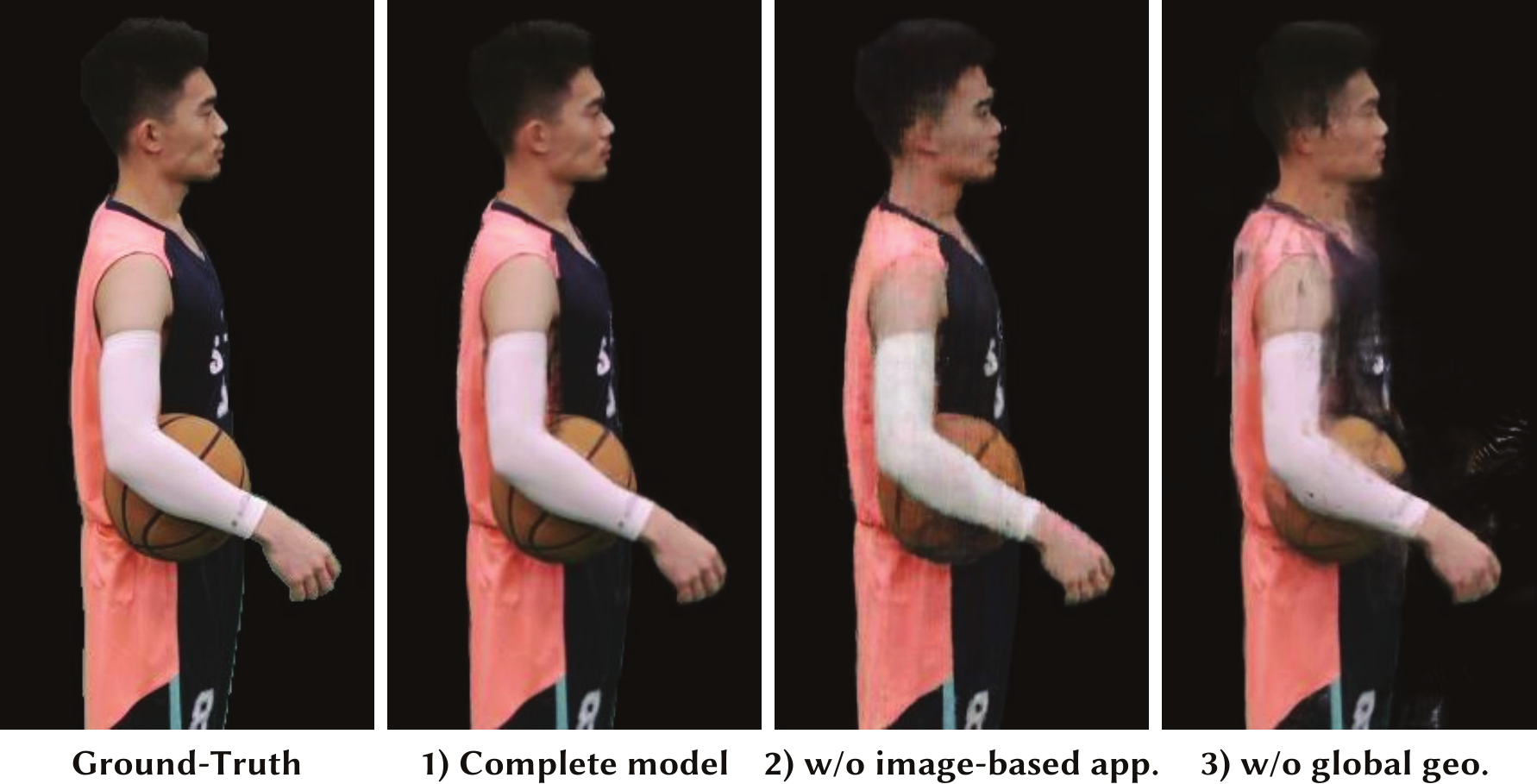}
    \caption{\textbf{Qualitative ablation study on the NHR dataset.} ``app.'' and ``geo.'' denote appearance and geometry, respectively.
    }
    \label{fig:ab_nhr}
\end{figure}

\paragraph{Ablations.}
We conduct qualitative and quantitative ablation studies on 2 sequences from NHR and DNA-Rendering datasets.
As shown in Table \ref{tab:ab_zju_nhr}, we first investigate the core components of our approach, i.e., the roles of the multi-view image-based appearance model and grid-based global geometry.
2) is analogous to K-Planes while 3) resembles IBRNet without the ray transformer proposed in IBRNet.
Quantitative results demonstrate that the absence of the multi-view image-based appearance model and global geometry would lead to extremely poor performance. This becomes more evident from the Fig. \ref{fig:ab_nhr}, where the lack of the multi-view image-based appearance model results in lost image detail, and the absence of grid-based \textit{global} geometry induces numerous floater artifacts.
Furthermore, we explore the effects of other components of our method, including the proposed efficient training strategy.
Initially, we find that not training the image-based appearance model, as in 4), results in poor performance. Not employing our specialized training strategy, as in 7), leads to more extended training periods (approximately an extra hour).
Subsequently, we discover that missing the joint training ($\numJointTraining = 0$) or not finetuning the image-feature network ($\numFinetuneTraining = \infty$) would result in slightly inferior outcomes.

\paragraph{Rendering time analysis.} 
Our rendering time consists of the time for the images to pass through the feature network and the time for volume rendering each ray.
On the ZJU-MoCap dataset, the time for the images to pass through the feature network is 1.26ms, and the time for rendering 512x512 rays is 11.27ms.
Without using the acceleration strategy, rendering 512x512 rays takes 449.1ms.
The final rendering times are 79.8FPS and 2.22FPS, respectively.
Note that after using the acceleration strategy, the rendering quality of our method has slightly decreased (0.09 in PSNR, 0.0004 in LPIPS), which we believe is acceptable.

\paragraph{Storage analysis.}
The storage of the proposed representation consists of the storage for the feature network, the spatial-temporal feature grids, the small MLPs, the binary field (for efficient rendering), and multiple videos. 
On the NHR dataset, the storage for the feature network is 161KB, the storage for the spatial-temporal feature grids is 82MB, the storage for the small MLPs is 154KB, the binary field is 2.2MB, and the storage for multiple (52 for NHR) lossless (pngs archived with zip format) image sequences is 300.85MB. 
By using the video compression techniques referenced in \cite{sullivan2012overview}, the image sequences can be compressed to 11.14MB with a PSNR error within 0.2 and an LPIPS error within 0.002.

\begin{table}
    \tabcolsep=0.1 cm
    \begin{center}
    \caption{\textbf{Quantitative ablation study on NHR and DNA-Rendering datasets.} ``Tt'' represents the training time, and its corresponding unit is hours. All the models are trained with the same training iterations. 
    Please refer to Sec. \ref{sec:ablation} for the detailed description.}
    \vspace{-2mm}
    \label{tab:ab_zju_nhr}
    \resizebox{\columnwidth}{!}{
    \begin{tabular}{l|ccc|ccc}
    \Xhline{3\arrayrulewidth}
     &  \multicolumn{3}{c|}{NHR} & \multicolumn{3}{c}{DNA-Rendering} \\
      & PSNR $\uparrow$  & LPIPS $\downarrow$  & Tt $\downarrow$ & PSNR $\uparrow$  & LPIPS $\downarrow$  & Tt $\downarrow$ \\
    \hline
    1) Complete model & 34.87 & 0.043 & 2.29 & 29.82 & 0.045 & 3.33  \\
    2) w/o \textit{image-based} appearance &  31.79 & 0.107 & 1.81 & 26.07 & 0.102 & 2.54  \\
    3) w/o \textit{global} geometry     &  24.71 & 0.152 & 2.81 & 27.44 & 0.100 & 4.04  \\
    \hline
    4) $\numJointTraining = 0$, \quad\, $\numFinetuneTraining = \infty$ & 30.44 & 0.051 & 2.21 &  28.25 & 0.047 & 3.23 \\
    5) $\numJointTraining = 0$, \quad\, $\numFinetuneTraining = 20$  & 34.30 & 0.044 & 2.25 & 29.63 & 0.047 & 3.29 \\
    6) $\numJointTraining = 5000$, $\numFinetuneTraining = \infty$ & 34.29 & 0.043 & 2.26 & 29.27 & 0.045 & 3.27 \\
    7) $\numJointTraining = \infty$, \quad $\numFinetuneTraining = \textbackslash$ &34.89 & 0.042 & 3.19 & 29.83 & 0.045 & 4.55 \\ 
    \Xhline{3\arrayrulewidth}
    \end{tabular}
    }
    \end{center}
    \vspace{-3mm}
\end{table}

\section{Conclusion and discussion}
\label{sec:conclusion}
This paper introduced Im4D, a novel hybrid scene representation for dynamic scenes that consists of a grid-based 4D geometry representation and a multi-view image-based appearance representation. 
The proposed method consistently demonstrates superior performance across various benchmarks for dynamic view synthesis including DyNeRF, NHR, ZJU-MoCap and DNA-Rendering datasets, achieving state-of-the-art rendering quality within a few hours of training. 
Furthermore, the proposed scene representation can be rendered in real-time and handle complex motions
within a diverse set of scenarios including outdoor challenging scenes as shown in our video, demonstrating its robustness.

This work still has some limitations. 
Although the proposed method significantly reduces the rendering artifacts of previous multi-view image-based rendering methods, it has the natural limitation that some regions may be occluded in the input views, which may lead to incorrect appearance prediction.
Future work could consider leveraging our consistent dynamic geometry for improved occlusion handling and source view selection to address this issue.
Another limitation is that our method cannot handle the monocular video as input, since our appearance representation requires multiple input views at the same moment.
\begin{acks}
    The authors would like to acknowledge support from NSFC (No. 62172364), Information Technology Center and State Key Lab of CAD\&CG, Zhejiang University.
\end{acks}

\newpage
\bibliographystyle{ACM-Reference-Format}
\bibliography{ref}


\begin{thebibliography}{96}


\ifx \showCODEN    \undefined \def \showCODEN     #1{\unskip}     \fi
\ifx \showDOI      \undefined \def \showDOI       #1{#1}\fi
\ifx \showISBNx    \undefined \def \showISBNx     #1{\unskip}     \fi
\ifx \showISBNxiii \undefined \def \showISBNxiii  #1{\unskip}     \fi
\ifx \showISSN     \undefined \def \showISSN      #1{\unskip}     \fi
\ifx \showLCCN     \undefined \def \showLCCN      #1{\unskip}     \fi
\ifx \shownote     \undefined \def \shownote      #1{#1}          \fi
\ifx \showarticletitle \undefined \def \showarticletitle #1{#1}   \fi
\ifx \showURL      \undefined \def \showURL       {\relax}        \fi
\providecommand\bibfield[2]{#2}
\providecommand\bibinfo[2]{#2}
\providecommand\natexlab[1]{#1}
\providecommand\showeprint[2][]{arXiv:#2}

\bibitem[Attal et~al\mbox{.}(2023)]%
        {attal2023hyperreel}
\bibfield{author}{\bibinfo{person}{Benjamin Attal}, \bibinfo{person}{Jia-Bin Huang}, \bibinfo{person}{Christian Richardt}, \bibinfo{person}{Michael Zollhoefer}, \bibinfo{person}{Johannes Kopf}, \bibinfo{person}{Matthew O'Toole}, {and} \bibinfo{person}{Changil Kim}.} \bibinfo{year}{2023}\natexlab{}.
\newblock \showarticletitle{HyperReel: High-Fidelity 6-DoF Video with Ray-Conditioned Sampling}.
\newblock \bibinfo{journal}{\emph{arXiv preprint arXiv:2301.02238}} (\bibinfo{year}{2023}).
\newblock


\bibitem[Attal et~al\mbox{.}(2022)]%
        {attal2022learning}
\bibfield{author}{\bibinfo{person}{Benjamin Attal}, \bibinfo{person}{Jia-Bin Huang}, \bibinfo{person}{Michael Zollh{\"o}fer}, \bibinfo{person}{Johannes Kopf}, {and} \bibinfo{person}{Changil Kim}.} \bibinfo{year}{2022}\natexlab{}.
\newblock \showarticletitle{Learning neural light fields with ray-space embedding}. In \bibinfo{booktitle}{\emph{Proceedings of the IEEE/CVF Conference on Computer Vision and Pattern Recognition}}. \bibinfo{pages}{19819--19829}.
\newblock


\bibitem[Barron et~al\mbox{.}(2021)]%
        {barron2021mip}
\bibfield{author}{\bibinfo{person}{Jonathan~T Barron}, \bibinfo{person}{Ben Mildenhall}, \bibinfo{person}{Dor Verbin}, \bibinfo{person}{Pratul~P Srinivasan}, {and} \bibinfo{person}{Peter Hedman}.} \bibinfo{year}{2021}\natexlab{}.
\newblock \showarticletitle{Mip-NeRF 360: Unbounded Anti-Aliased Neural Radiance Fields}.
\newblock \bibinfo{journal}{\emph{arXiv}} (\bibinfo{year}{2021}).
\newblock


\bibitem[Barron et~al\mbox{.}(2022)]%
        {barron2022mip}
\bibfield{author}{\bibinfo{person}{Jonathan~T Barron}, \bibinfo{person}{Ben Mildenhall}, \bibinfo{person}{Dor Verbin}, \bibinfo{person}{Pratul~P Srinivasan}, {and} \bibinfo{person}{Peter Hedman}.} \bibinfo{year}{2022}\natexlab{}.
\newblock \showarticletitle{Mip-nerf 360: Unbounded anti-aliased neural radiance fields}. In \bibinfo{booktitle}{\emph{Proceedings of the IEEE/CVF Conference on Computer Vision and Pattern Recognition}}. \bibinfo{pages}{5470--5479}.
\newblock


\bibitem[Barron et~al\mbox{.}(2023)]%
        {barron2023zip}
\bibfield{author}{\bibinfo{person}{Jonathan~T Barron}, \bibinfo{person}{Ben Mildenhall}, \bibinfo{person}{Dor Verbin}, \bibinfo{person}{Pratul~P Srinivasan}, {and} \bibinfo{person}{Peter Hedman}.} \bibinfo{year}{2023}\natexlab{}.
\newblock \showarticletitle{Zip-NeRF: Anti-Aliased Grid-Based Neural Radiance Fields}.
\newblock \bibinfo{journal}{\emph{arXiv preprint arXiv:2304.06706}} (\bibinfo{year}{2023}).
\newblock


\bibitem[Bergman et~al\mbox{.}(2021)]%
        {bergman2021fast}
\bibfield{author}{\bibinfo{person}{Alexander Bergman}, \bibinfo{person}{Petr Kellnhofer}, {and} \bibinfo{person}{Gordon Wetzstein}.} \bibinfo{year}{2021}\natexlab{}.
\newblock \showarticletitle{Fast training of neural lumigraph representations using meta learning}.
\newblock \bibinfo{journal}{\emph{Advances in Neural Information Processing Systems}}  \bibinfo{volume}{34} (\bibinfo{year}{2021}), \bibinfo{pages}{172--186}.
\newblock


\bibitem[Buehler et~al\mbox{.}(2001)]%
        {chris2001unstructured}
\bibfield{author}{\bibinfo{person}{Chris Buehler}, \bibinfo{person}{Michael Bosse}, \bibinfo{person}{Leonard McMillan}, \bibinfo{person}{Steven Gortler}, {and} \bibinfo{person}{Michael Cohen}.} \bibinfo{year}{2001}\natexlab{}.
\newblock \showarticletitle{Unstructured Lumigraph Rendering}. In \bibinfo{booktitle}{\emph{Proceedings of the 28th Annual Conference on Computer Graphics and Interactive Techniques}} \emph{(\bibinfo{series}{SIGGRAPH '01})}. \bibinfo{publisher}{Association for Computing Machinery}, \bibinfo{address}{New York, NY, USA}, \bibinfo{pages}{425–432}.
\newblock
\showISBNx{158113374X}
\urldef\tempurl%
\url{https://doi.org/10.1145/383259.383309}
\showDOI{\tempurl}


\bibitem[Cao and Johnson(2023)]%
        {cao2023hexplane}
\bibfield{author}{\bibinfo{person}{Ang Cao} {and} \bibinfo{person}{Justin Johnson}.} \bibinfo{year}{2023}\natexlab{}.
\newblock \showarticletitle{Hexplane: A fast representation for dynamic scenes}.
\newblock \bibinfo{journal}{\emph{arXiv}} (\bibinfo{year}{2023}).
\newblock


\bibitem[Chan et~al\mbox{.}(2022)]%
        {chan2022efficient}
\bibfield{author}{\bibinfo{person}{Eric~R Chan}, \bibinfo{person}{Connor~Z Lin}, \bibinfo{person}{Matthew~A Chan}, \bibinfo{person}{Koki Nagano}, \bibinfo{person}{Boxiao Pan}, \bibinfo{person}{Shalini De~Mello}, \bibinfo{person}{Orazio Gallo}, \bibinfo{person}{Leonidas~J Guibas}, \bibinfo{person}{Jonathan Tremblay}, \bibinfo{person}{Sameh Khamis}, {et~al\mbox{.}}} \bibinfo{year}{2022}\natexlab{}.
\newblock \showarticletitle{Efficient geometry-aware 3D generative adversarial networks}. In \bibinfo{booktitle}{\emph{CVPR}}.
\newblock


\bibitem[Chaurasia et~al\mbox{.}(2013)]%
        {chaurasia2013depth}
\bibfield{author}{\bibinfo{person}{Gaurav Chaurasia}, \bibinfo{person}{Sylvain Duchene}, \bibinfo{person}{Olga Sorkine-Hornung}, {and} \bibinfo{person}{George Drettakis}.} \bibinfo{year}{2013}\natexlab{}.
\newblock \showarticletitle{Depth synthesis and local warps for plausible image-based navigation}.
\newblock \bibinfo{journal}{\emph{ACM TOG}} (\bibinfo{year}{2013}).
\newblock


\bibitem[Chen et~al\mbox{.}(2022b)]%
        {chen2022tensorf}
\bibfield{author}{\bibinfo{person}{Anpei Chen}, \bibinfo{person}{Zexiang Xu}, \bibinfo{person}{Andreas Geiger}, \bibinfo{person}{Jingyi Yu}, {and} \bibinfo{person}{Hao Su}.} \bibinfo{year}{2022}\natexlab{b}.
\newblock \showarticletitle{TensoRF: Tensorial Radiance Fields}.
\newblock \bibinfo{journal}{\emph{arXiv}} (\bibinfo{year}{2022}).
\newblock


\bibitem[Chen et~al\mbox{.}(2021)]%
        {chen2021mvsnerf}
\bibfield{author}{\bibinfo{person}{Anpei Chen}, \bibinfo{person}{Zexiang Xu}, \bibinfo{person}{Fuqiang Zhao}, \bibinfo{person}{Xiaoshuai Zhang}, \bibinfo{person}{Fanbo Xiang}, \bibinfo{person}{Jingyi Yu}, {and} \bibinfo{person}{Hao Su}.} \bibinfo{year}{2021}\natexlab{}.
\newblock \showarticletitle{MVSNeRF: Fast Generalizable Radiance Field Reconstruction From Multi-View Stereo}. In \bibinfo{booktitle}{\emph{ICCV}}.
\newblock


\bibitem[Chen et~al\mbox{.}(2022a)]%
        {chen2022mobilenerf}
\bibfield{author}{\bibinfo{person}{Zhiqin Chen}, \bibinfo{person}{Thomas Funkhouser}, \bibinfo{person}{Peter Hedman}, {and} \bibinfo{person}{Andrea Tagliasacchi}.} \bibinfo{year}{2022}\natexlab{a}.
\newblock \showarticletitle{Mobilenerf: Exploiting the polygon rasterization pipeline for efficient neural field rendering on mobile architectures}.
\newblock \bibinfo{journal}{\emph{arXiv preprint arXiv:2208.00277}} (\bibinfo{year}{2022}).
\newblock


\bibitem[Cheng et~al\mbox{.}(2023)]%
        {renbody}
\bibfield{author}{\bibinfo{person}{Wei Cheng}, \bibinfo{person}{Ruixiang Chen}, \bibinfo{person}{Wanqi Yin}, \bibinfo{person}{Siming Fan}, \bibinfo{person}{Keyu Chen}, \bibinfo{person}{Honglin He}, \bibinfo{person}{Huiwen Luo}, \bibinfo{person}{Zhongang Cai}, \bibinfo{person}{Jingbo Wang}, \bibinfo{person}{Yang Gao}, \bibinfo{person}{Zhengming Yu}, \bibinfo{person}{Zhengyu Lin}, \bibinfo{person}{Daxuan Ren}, \bibinfo{person}{Lei Yang}, \bibinfo{person}{Ziwei Liu}, \bibinfo{person}{Chen~Change Loy}, \bibinfo{person}{Chen Qian}, \bibinfo{person}{Wayne Wu}, \bibinfo{person}{Dahua Lin}, \bibinfo{person}{Bo Dai}, {and} \bibinfo{person}{Kwan-Yee Lin}.} \bibinfo{year}{2023}\natexlab{}.
\newblock \showarticletitle{DNA-Rendering: A Diverse Neural Actor Repository for High-Fidelity Human-centric Rendering}.
\newblock \bibinfo{journal}{\emph{ICCV}} (\bibinfo{year}{2023}).
\newblock


\bibitem[Chibane et~al\mbox{.}(2021)]%
        {chibane2021stereo}
\bibfield{author}{\bibinfo{person}{Julian Chibane}, \bibinfo{person}{Aayush Bansal}, \bibinfo{person}{Verica Lazova}, {and} \bibinfo{person}{Gerard Pons-Moll}.} \bibinfo{year}{2021}\natexlab{}.
\newblock \showarticletitle{Stereo Radiance Fields (SRF): Learning View Synthesis for Sparse Views of Novel Scenes}. In \bibinfo{booktitle}{\emph{CVPR}}.
\newblock


\bibitem[Collet et~al\mbox{.}(2015)]%
        {collet2015high}
\bibfield{author}{\bibinfo{person}{Alvaro Collet}, \bibinfo{person}{Ming Chuang}, \bibinfo{person}{Pat Sweeney}, \bibinfo{person}{Don Gillett}, \bibinfo{person}{Dennis Evseev}, \bibinfo{person}{David Calabrese}, \bibinfo{person}{Hugues Hoppe}, \bibinfo{person}{Adam Kirk}, {and} \bibinfo{person}{Steve Sullivan}.} \bibinfo{year}{2015}\natexlab{}.
\newblock \showarticletitle{High-quality streamable free-viewpoint video}.
\newblock \bibinfo{journal}{\emph{ACM TOG}} (\bibinfo{year}{2015}).
\newblock


\bibitem[Davis et~al\mbox{.}(2012)]%
        {davis2012unstructured}
\bibfield{author}{\bibinfo{person}{Abe Davis}, \bibinfo{person}{Marc Levoy}, {and} \bibinfo{person}{Fredo Durand}.} \bibinfo{year}{2012}\natexlab{}.
\newblock \showarticletitle{Unstructured light fields}. In \bibinfo{booktitle}{\emph{Eurographics}}.
\newblock


\bibitem[Dou et~al\mbox{.}(2016)]%
        {dou2016fusion4d}
\bibfield{author}{\bibinfo{person}{Mingsong Dou}, \bibinfo{person}{Sameh Khamis}, \bibinfo{person}{Yury Degtyarev}, \bibinfo{person}{Philip Davidson}, \bibinfo{person}{Sean~Ryan Fanello}, \bibinfo{person}{Adarsh Kowdle}, \bibinfo{person}{Sergio~Orts Escolano}, \bibinfo{person}{Christoph Rhemann}, \bibinfo{person}{David Kim}, \bibinfo{person}{Jonathan Taylor}, {et~al\mbox{.}}} \bibinfo{year}{2016}\natexlab{}.
\newblock \showarticletitle{Fusion4d: Real-time performance capture of challenging scenes}.
\newblock \bibinfo{journal}{\emph{ACM TOG}} (\bibinfo{year}{2016}).
\newblock


\bibitem[Du et~al\mbox{.}(2021)]%
        {du2021nerflow}
\bibfield{author}{\bibinfo{person}{Yilun Du}, \bibinfo{person}{Yinan Zhang}, \bibinfo{person}{Hong-Xing Yu}, \bibinfo{person}{Joshua~B. Tenenbaum}, {and} \bibinfo{person}{Jiajun Wu}.} \bibinfo{year}{2021}\natexlab{}.
\newblock \showarticletitle{Neural Radiance Flow for 4D View Synthesis and Video Processing}. In \bibinfo{booktitle}{\emph{ICCV}}.
\newblock


\bibitem[Fang et~al\mbox{.}(2022)]%
        {fang2022fast}
\bibfield{author}{\bibinfo{person}{Jiemin Fang}, \bibinfo{person}{Taoran Yi}, \bibinfo{person}{Xinggang Wang}, \bibinfo{person}{Lingxi Xie}, \bibinfo{person}{Xiaopeng Zhang}, \bibinfo{person}{Wenyu Liu}, \bibinfo{person}{Matthias Nie{\ss}ner}, {and} \bibinfo{person}{Qi Tian}.} \bibinfo{year}{2022}\natexlab{}.
\newblock \showarticletitle{Fast dynamic radiance fields with time-aware neural voxels}. In \bibinfo{booktitle}{\emph{SIGGRAPH Asia 2022 Conference Papers}}. \bibinfo{pages}{1--9}.
\newblock


\bibitem[Flynn et~al\mbox{.}(2016)]%
        {flynn2016deepstereo}
\bibfield{author}{\bibinfo{person}{John Flynn}, \bibinfo{person}{Ivan Neulander}, \bibinfo{person}{James Philbin}, {and} \bibinfo{person}{Noah Snavely}.} \bibinfo{year}{2016}\natexlab{}.
\newblock \showarticletitle{DeepStereo: Learning to Predict New Views From the World's Imagery}. In \bibinfo{booktitle}{\emph{CVPR}}.
\newblock


\bibitem[Fridovich-Keil et~al\mbox{.}(2023)]%
        {fridovich2023k}
\bibfield{author}{\bibinfo{person}{Sara Fridovich-Keil}, \bibinfo{person}{Giacomo Meanti}, \bibinfo{person}{Frederik Warburg}, \bibinfo{person}{Benjamin Recht}, {and} \bibinfo{person}{Angjoo Kanazawa}.} \bibinfo{year}{2023}\natexlab{}.
\newblock \showarticletitle{K-planes: Explicit radiance fields in space, time, and appearance}.
\newblock \bibinfo{journal}{\emph{arXiv}} (\bibinfo{year}{2023}).
\newblock


\bibitem[Gan et~al\mbox{.}(2022)]%
        {gan2022v4d}
\bibfield{author}{\bibinfo{person}{Wanshui Gan}, \bibinfo{person}{Hongbin Xu}, \bibinfo{person}{Yi Huang}, \bibinfo{person}{Shifeng Chen}, {and} \bibinfo{person}{Naoto Yokoya}.} \bibinfo{year}{2022}\natexlab{}.
\newblock \showarticletitle{V4d: Voxel for 4d novel view synthesis}.
\newblock \bibinfo{journal}{\emph{arXiv preprint arXiv:2205.14332}} (\bibinfo{year}{2022}).
\newblock


\bibitem[Gao et~al\mbox{.}(2021)]%
        {gao2021dynamic}
\bibfield{author}{\bibinfo{person}{Chen Gao}, \bibinfo{person}{Ayush Saraf}, \bibinfo{person}{Johannes Kopf}, {and} \bibinfo{person}{Jia-Bin Huang}.} \bibinfo{year}{2021}\natexlab{}.
\newblock \showarticletitle{Dynamic view synthesis from dynamic monocular video}. In \bibinfo{booktitle}{\emph{Proceedings of the IEEE/CVF International Conference on Computer Vision}}. \bibinfo{pages}{5712--5721}.
\newblock


\bibitem[Garbin et~al\mbox{.}(2021)]%
        {garbin2021fastnerf}
\bibfield{author}{\bibinfo{person}{Stephan~J. Garbin}, \bibinfo{person}{Marek Kowalski}, \bibinfo{person}{Matthew Johnson}, \bibinfo{person}{Jamie Shotton}, {and} \bibinfo{person}{Julien Valentin}.} \bibinfo{year}{2021}\natexlab{}.
\newblock \showarticletitle{FastNeRF: High-Fidelity Neural Rendering at 200FPS}. In \bibinfo{booktitle}{\emph{ICCV}}.
\newblock


\bibitem[Gortler et~al\mbox{.}(1996)]%
        {gortler1996lumigraph}
\bibfield{author}{\bibinfo{person}{Steven~J Gortler}, \bibinfo{person}{Radek Grzeszczuk}, \bibinfo{person}{Richard Szeliski}, {and} \bibinfo{person}{Michael~F Cohen}.} \bibinfo{year}{1996}\natexlab{}.
\newblock \showarticletitle{The lumigraph}. In \bibinfo{booktitle}{\emph{SIGGRAPH}}.
\newblock


\bibitem[Hedman et~al\mbox{.}(2018)]%
        {hedman2018deep}
\bibfield{author}{\bibinfo{person}{Peter Hedman}, \bibinfo{person}{Julien Philip}, \bibinfo{person}{True Price}, \bibinfo{person}{Jan-Michael Frahm}, \bibinfo{person}{George Drettakis}, {and} \bibinfo{person}{Gabriel Brostow}.} \bibinfo{year}{2018}\natexlab{}.
\newblock \showarticletitle{Deep blending for free-viewpoint image-based rendering}.
\newblock \bibinfo{journal}{\emph{ACM TOG}} (\bibinfo{year}{2018}).
\newblock


\bibitem[Hedman et~al\mbox{.}(2021)]%
        {hedman2021baking}
\bibfield{author}{\bibinfo{person}{Peter Hedman}, \bibinfo{person}{Pratul~P. Srinivasan}, \bibinfo{person}{Ben Mildenhall}, \bibinfo{person}{Jonathan~T. Barron}, {and} \bibinfo{person}{Paul Debevec}.} \bibinfo{year}{2021}\natexlab{}.
\newblock \showarticletitle{Baking Neural Radiance Fields for Real-Time View Synthesis}. In \bibinfo{booktitle}{\emph{ICCV}}.
\newblock


\bibitem[I{\c{s}}{\i}k et~al\mbox{.}(2023)]%
        {icsik2023humanrf}
\bibfield{author}{\bibinfo{person}{Mustafa I{\c{s}}{\i}k}, \bibinfo{person}{Martin R{\"u}nz}, \bibinfo{person}{Markos Georgopoulos}, \bibinfo{person}{Taras Khakhulin}, \bibinfo{person}{Jonathan Starck}, \bibinfo{person}{Lourdes Agapito}, {and} \bibinfo{person}{Matthias Nie{\ss}ner}.} \bibinfo{year}{2023}\natexlab{}.
\newblock \showarticletitle{HumanRF: High-Fidelity Neural Radiance Fields for Humans in Motion}.
\newblock \bibinfo{journal}{\emph{arXiv preprint arXiv:2305.06356}} (\bibinfo{year}{2023}).
\newblock


\bibitem[Jiang et~al\mbox{.}(2020)]%
        {jiang2020sdfdiff}
\bibfield{author}{\bibinfo{person}{Yue Jiang}, \bibinfo{person}{Dantong Ji}, \bibinfo{person}{Zhizhong Han}, {and} \bibinfo{person}{Matthias Zwicker}.} \bibinfo{year}{2020}\natexlab{}.
\newblock \showarticletitle{Sdfdiff: Differentiable rendering of signed distance fields for 3d shape optimization}. In \bibinfo{booktitle}{\emph{CVPR}}.
\newblock


\bibitem[Johari et~al\mbox{.}(2022)]%
        {johari2021geonerf}
\bibfield{author}{\bibinfo{person}{Mohammad~Mahdi Johari}, \bibinfo{person}{Yann Lepoittevin}, {and} \bibinfo{person}{Fran{\c{c}}ois Fleuret}.} \bibinfo{year}{2022}\natexlab{}.
\newblock \showarticletitle{GeoNeRF: Generalizing NeRF with Geometry Priors}.
\newblock \bibinfo{journal}{\emph{CVPR}} (\bibinfo{year}{2022}).
\newblock


\bibitem[Kalantari et~al\mbox{.}(2016)]%
        {kalantari2016learning}
\bibfield{author}{\bibinfo{person}{Nima~Khademi Kalantari}, \bibinfo{person}{Ting-Chun Wang}, {and} \bibinfo{person}{Ravi Ramamoorthi}.} \bibinfo{year}{2016}\natexlab{}.
\newblock \showarticletitle{Learning-based view synthesis for light field cameras}.
\newblock \bibinfo{journal}{\emph{ACM TOG}} (\bibinfo{year}{2016}).
\newblock


\bibitem[Kellnhofer et~al\mbox{.}(2021)]%
        {kellnhofer2021neural}
\bibfield{author}{\bibinfo{person}{Petr Kellnhofer}, \bibinfo{person}{Lars~C Jebe}, \bibinfo{person}{Andrew Jones}, \bibinfo{person}{Ryan Spicer}, \bibinfo{person}{Kari Pulli}, {and} \bibinfo{person}{Gordon Wetzstein}.} \bibinfo{year}{2021}\natexlab{}.
\newblock \showarticletitle{Neural lumigraph rendering}. In \bibinfo{booktitle}{\emph{Proceedings of the IEEE/CVF Conference on Computer Vision and Pattern Recognition}}. \bibinfo{pages}{4287--4297}.
\newblock


\bibitem[Kingma and Ba(2015)]%
        {kingma2014adam}
\bibfield{author}{\bibinfo{person}{Diederik~P. Kingma} {and} \bibinfo{person}{Jimmy Ba}.} \bibinfo{year}{2015}\natexlab{}.
\newblock \showarticletitle{Adam: {A} Method for Stochastic Optimization}. In \bibinfo{booktitle}{\emph{ICLR}}.
\newblock
\urldef\tempurl%
\url{http://arxiv.org/abs/1412.6980}
\showURL{%
\tempurl}


\bibitem[Kopanas et~al\mbox{.}(2021)]%
        {kopanas2021point}
\bibfield{author}{\bibinfo{person}{Georgios Kopanas}, \bibinfo{person}{Julien Philip}, \bibinfo{person}{Thomas Leimk{\"u}hler}, {and} \bibinfo{person}{George Drettakis}.} \bibinfo{year}{2021}\natexlab{}.
\newblock \showarticletitle{Point-Based Neural Rendering with Per-View Optimization}. In \bibinfo{booktitle}{\emph{Computer Graphics Forum}}, Vol.~\bibinfo{volume}{40}. Wiley Online Library, \bibinfo{pages}{29--43}.
\newblock


\bibitem[Levoy and Hanrahan(1996)]%
        {levoy1996light}
\bibfield{author}{\bibinfo{person}{Marc Levoy} {and} \bibinfo{person}{Pat Hanrahan}.} \bibinfo{year}{1996}\natexlab{}.
\newblock \showarticletitle{Light field rendering}. In \bibinfo{booktitle}{\emph{SIGGRAPH}}.
\newblock


\bibitem[Li et~al\mbox{.}(2022a)]%
        {li2022streaming}
\bibfield{author}{\bibinfo{person}{Lingzhi Li}, \bibinfo{person}{Zhen Shen}, \bibinfo{person}{Li Shen}, \bibinfo{person}{Ping Tan}, {et~al\mbox{.}}} \bibinfo{year}{2022}\natexlab{a}.
\newblock \showarticletitle{Streaming Radiance Fields for 3D Video Synthesis}. In \bibinfo{booktitle}{\emph{Advances in Neural Information Processing Systems}}.
\newblock


\bibitem[Li et~al\mbox{.}(2023)]%
        {li2023nerfacc}
\bibfield{author}{\bibinfo{person}{Ruilong Li}, \bibinfo{person}{Hang Gao}, \bibinfo{person}{Matthew Tancik}, {and} \bibinfo{person}{Angjoo Kanazawa}.} \bibinfo{year}{2023}\natexlab{}.
\newblock \showarticletitle{NerfAcc: Efficient Sampling Accelerates NeRFs.}
\newblock \bibinfo{journal}{\emph{arXiv preprint arXiv:2305.04966}} (\bibinfo{year}{2023}).
\newblock


\bibitem[Li et~al\mbox{.}(2021b)]%
        {li2021neural}
\bibfield{author}{\bibinfo{person}{Tianye Li}, \bibinfo{person}{Mira Slavcheva}, \bibinfo{person}{Michael Zollhoefer}, \bibinfo{person}{Simon Green}, \bibinfo{person}{Christoph Lassner}, \bibinfo{person}{Changil Kim}, \bibinfo{person}{Tanner Schmidt}, \bibinfo{person}{Steven Lovegrove}, \bibinfo{person}{Michael Goesele}, {and} \bibinfo{person}{Zhaoyang Lv}.} \bibinfo{year}{2021}\natexlab{b}.
\newblock \showarticletitle{Neural 3d video synthesis}.
\newblock \bibinfo{journal}{\emph{arXiv preprint arXiv:2103.02597}} (\bibinfo{year}{2021}).
\newblock


\bibitem[Li et~al\mbox{.}(2022b)]%
        {li2021neural3d}
\bibfield{author}{\bibinfo{person}{Tianye Li}, \bibinfo{person}{Mira Slavcheva}, \bibinfo{person}{Michael Zollhoefer}, \bibinfo{person}{Simon Green}, \bibinfo{person}{Christoph Lassner}, \bibinfo{person}{Changil Kim}, \bibinfo{person}{Tanner Schmidt}, \bibinfo{person}{Steven Lovegrove}, \bibinfo{person}{Michael Goesele}, {and} \bibinfo{person}{Zhaoyang Lv}.} \bibinfo{year}{2022}\natexlab{b}.
\newblock \showarticletitle{Neural 3d video synthesis}.
\newblock \bibinfo{journal}{\emph{CVPR}} (\bibinfo{year}{2022}).
\newblock


\bibitem[Li et~al\mbox{.}(2021a)]%
        {li2020neural}
\bibfield{author}{\bibinfo{person}{Zhengqi Li}, \bibinfo{person}{Simon Niklaus}, \bibinfo{person}{Noah Snavely}, {and} \bibinfo{person}{Oliver Wang}.} \bibinfo{year}{2021}\natexlab{a}.
\newblock \showarticletitle{Neural Scene Flow Fields for Space-Time View Synthesis of Dynamic Scenes}. In \bibinfo{booktitle}{\emph{CVPR}}.
\newblock


\bibitem[Li et~al\mbox{.}(2020)]%
        {li2020crowdsampling}
\bibfield{author}{\bibinfo{person}{Zhengqi Li}, \bibinfo{person}{Wenqi Xian}, \bibinfo{person}{Abe Davis}, {and} \bibinfo{person}{Noah Snavely}.} \bibinfo{year}{2020}\natexlab{}.
\newblock \showarticletitle{Crowdsampling the plenoptic function}. In \bibinfo{booktitle}{\emph{ECCV}}.
\newblock


\bibitem[Lin et~al\mbox{.}(2022)]%
        {lin2022enerf}
\bibfield{author}{\bibinfo{person}{Haotong Lin}, \bibinfo{person}{Sida Peng}, \bibinfo{person}{Zhen Xu}, \bibinfo{person}{Yunzhi Yan}, \bibinfo{person}{Qing Shuai}, \bibinfo{person}{Hujun Bao}, {and} \bibinfo{person}{Xiaowei Zhou}.} \bibinfo{year}{2022}\natexlab{}.
\newblock \showarticletitle{Efficient Neural Radiance Fields for Interactive Free-viewpoint Video}. In \bibinfo{booktitle}{\emph{SIGGRAPH Asia Conference Proceedings}}.
\newblock


\bibitem[Lin et~al\mbox{.}(2023)]%
        {lin2023neural}
\bibfield{author}{\bibinfo{person}{Haotong Lin}, \bibinfo{person}{Qianqian Wang}, \bibinfo{person}{Ruojin Cai}, \bibinfo{person}{Sida Peng}, \bibinfo{person}{Hadar Averbuch-Elor}, \bibinfo{person}{Xiaowei Zhou}, {and} \bibinfo{person}{Noah Snavely}.} \bibinfo{year}{2023}\natexlab{}.
\newblock \showarticletitle{Neural Scene Chronology}. In \bibinfo{booktitle}{\emph{Proceedings of the IEEE/CVF Conference on Computer Vision and Pattern Recognition}}. \bibinfo{pages}{20752--20761}.
\newblock


\bibitem[Liu et~al\mbox{.}(2020)]%
        {liu2020neural}
\bibfield{author}{\bibinfo{person}{Lingjie Liu}, \bibinfo{person}{Jiatao Gu}, \bibinfo{person}{Kyaw Zaw~Lin}, \bibinfo{person}{Tat-Seng Chua}, {and} \bibinfo{person}{Christian Theobalt}.} \bibinfo{year}{2020}\natexlab{}.
\newblock \showarticletitle{Neural Sparse Voxel Fields}. In \bibinfo{booktitle}{\emph{NeurIPS}}.
\newblock
\urldef\tempurl%
\url{https://proceedings.neurips.cc/paper/2020/file/b4b758962f17808746e9bb832a6fa4b8-Paper.pdf}
\showURL{%
\tempurl}


\bibitem[Liu et~al\mbox{.}(2019b)]%
        {liu2019neural}
\bibfield{author}{\bibinfo{person}{Lingjie Liu}, \bibinfo{person}{Weipeng Xu}, \bibinfo{person}{Michael Zollhoefer}, \bibinfo{person}{Hyeongwoo Kim}, \bibinfo{person}{Florian Bernard}, \bibinfo{person}{Marc Habermann}, \bibinfo{person}{Wenping Wang}, {and} \bibinfo{person}{Christian Theobalt}.} \bibinfo{year}{2019}\natexlab{b}.
\newblock \showarticletitle{Neural rendering and reenactment of human actor videos}.
\newblock \bibinfo{journal}{\emph{ACM TOG}} (\bibinfo{year}{2019}).
\newblock


\bibitem[Liu et~al\mbox{.}(2019a)]%
        {liu2019learning}
\bibfield{author}{\bibinfo{person}{Shichen Liu}, \bibinfo{person}{Shunsuke Saito}, \bibinfo{person}{Weikai Chen}, {and} \bibinfo{person}{Hao Li}.} \bibinfo{year}{2019}\natexlab{a}.
\newblock \showarticletitle{Learning to infer implicit surfaces without 3d supervision}.
\newblock \bibinfo{journal}{\emph{NeurIPS}} (\bibinfo{year}{2019}).
\newblock


\bibitem[Liu et~al\mbox{.}(2021)]%
        {liu2021neural}
\bibfield{author}{\bibinfo{person}{Yuan Liu}, \bibinfo{person}{Sida Peng}, \bibinfo{person}{Lingjie Liu}, \bibinfo{person}{Qianqian Wang}, \bibinfo{person}{Peng Wang}, \bibinfo{person}{Christian Theobalt}, \bibinfo{person}{Xiaowei Zhou}, {and} \bibinfo{person}{Wenping Wang}.} \bibinfo{year}{2021}\natexlab{}.
\newblock \showarticletitle{Neural Rays for Occlusion-aware Image-based Rendering}.
\newblock \bibinfo{journal}{\emph{arXiv}} (\bibinfo{year}{2021}).
\newblock


\bibitem[Lombardi et~al\mbox{.}(2019)]%
        {lombardi2019neural}
\bibfield{author}{\bibinfo{person}{Stephen Lombardi}, \bibinfo{person}{Tomas Simon}, \bibinfo{person}{Jason Saragih}, \bibinfo{person}{Gabriel Schwartz}, \bibinfo{person}{Andreas Lehrmann}, {and} \bibinfo{person}{Yaser Sheikh}.} \bibinfo{year}{2019}\natexlab{}.
\newblock \showarticletitle{Neural volumes: Learning dynamic renderable volumes from images}. In \bibinfo{booktitle}{\emph{SIGGRAPH}}.
\newblock


\bibitem[Lombardi et~al\mbox{.}(2021)]%
        {lombardi2021mixture}
\bibfield{author}{\bibinfo{person}{Stephen Lombardi}, \bibinfo{person}{Tomas Simon}, \bibinfo{person}{Gabriel Schwartz}, \bibinfo{person}{Michael Zollhoefer}, \bibinfo{person}{Yaser Sheikh}, {and} \bibinfo{person}{Jason Saragih}.} \bibinfo{year}{2021}\natexlab{}.
\newblock \showarticletitle{Mixture of volumetric primitives for efficient neural rendering}.
\newblock \bibinfo{journal}{\emph{ACM Transactions on Graphics (TOG)}} \bibinfo{volume}{40}, \bibinfo{number}{4} (\bibinfo{year}{2021}), \bibinfo{pages}{1--13}.
\newblock


\bibitem[Mildenhall et~al\mbox{.}(2019)]%
        {mildenhall2019local}
\bibfield{author}{\bibinfo{person}{Ben Mildenhall}, \bibinfo{person}{Pratul~P Srinivasan}, \bibinfo{person}{Rodrigo Ortiz-Cayon}, \bibinfo{person}{Nima~Khademi Kalantari}, \bibinfo{person}{Ravi Ramamoorthi}, \bibinfo{person}{Ren Ng}, {and} \bibinfo{person}{Abhishek Kar}.} \bibinfo{year}{2019}\natexlab{}.
\newblock \showarticletitle{Local light field fusion: Practical view synthesis with prescriptive sampling guidelines}.
\newblock \bibinfo{journal}{\emph{ACM TOG}} (\bibinfo{year}{2019}).
\newblock


\bibitem[Mildenhall et~al\mbox{.}(2020)]%
        {mildenhall2020nerf}
\bibfield{author}{\bibinfo{person}{Ben Mildenhall}, \bibinfo{person}{Pratul~P Srinivasan}, \bibinfo{person}{Matthew Tancik}, \bibinfo{person}{Jonathan~T Barron}, \bibinfo{person}{Ravi Ramamoorthi}, {and} \bibinfo{person}{Ren Ng}.} \bibinfo{year}{2020}\natexlab{}.
\newblock \showarticletitle{Nerf: Representing scenes as neural radiance fields for view synthesis}. In \bibinfo{booktitle}{\emph{ECCV}}.
\newblock


\bibitem[M{\"u}ller et~al\mbox{.}(2022)]%
        {muller2022instant}
\bibfield{author}{\bibinfo{person}{Thomas M{\"u}ller}, \bibinfo{person}{Alex Evans}, \bibinfo{person}{Christoph Schied}, {and} \bibinfo{person}{Alexander Keller}.} \bibinfo{year}{2022}\natexlab{}.
\newblock \showarticletitle{Instant Neural Graphics Primitives with a Multiresolution Hash Encoding}.
\newblock \bibinfo{journal}{\emph{SIGGRAPH}} (\bibinfo{year}{2022}).
\newblock


\bibitem[Neff et~al\mbox{.}(2021)]%
        {neff2021donerf}
\bibfield{author}{\bibinfo{person}{Thomas Neff}, \bibinfo{person}{Pascal Stadlbauer}, \bibinfo{person}{Mathias Parger}, \bibinfo{person}{Andreas Kurz}, \bibinfo{person}{Joerg~H Mueller}, \bibinfo{person}{Chakravarty R~Alla Chaitanya}, \bibinfo{person}{Anton Kaplanyan}, {and} \bibinfo{person}{Markus Steinberger}.} \bibinfo{year}{2021}\natexlab{}.
\newblock \showarticletitle{DONeRF: Towards Real-Time Rendering of Compact Neural Radiance Fields using Depth Oracle Networks}. In \bibinfo{booktitle}{\emph{EGSR}}.
\newblock


\bibitem[Newcombe et~al\mbox{.}(2015)]%
        {newcombe2015dynamicfusion}
\bibfield{author}{\bibinfo{person}{Richard~A Newcombe}, \bibinfo{person}{Dieter Fox}, {and} \bibinfo{person}{Steven~M Seitz}.} \bibinfo{year}{2015}\natexlab{}.
\newblock \showarticletitle{Dynamicfusion: Reconstruction and tracking of non-rigid scenes in real-time}. In \bibinfo{booktitle}{\emph{CVPR}}.
\newblock


\bibitem[Orts-Escolano et~al\mbox{.}(2016)]%
        {orts2016holoportation}
\bibfield{author}{\bibinfo{person}{Sergio Orts-Escolano}, \bibinfo{person}{Christoph Rhemann}, \bibinfo{person}{Sean Fanello}, \bibinfo{person}{Wayne Chang}, \bibinfo{person}{Adarsh Kowdle}, \bibinfo{person}{Yury Degtyarev}, \bibinfo{person}{David Kim}, \bibinfo{person}{Philip~L Davidson}, \bibinfo{person}{Sameh Khamis}, \bibinfo{person}{Mingsong Dou}, {et~al\mbox{.}}} \bibinfo{year}{2016}\natexlab{}.
\newblock \showarticletitle{Holoportation: Virtual 3d teleportation in real-time}. In \bibinfo{booktitle}{\emph{UIST}}.
\newblock


\bibitem[Park et~al\mbox{.}(2021a)]%
        {park2021nerfies}
\bibfield{author}{\bibinfo{person}{Keunhong Park}, \bibinfo{person}{Utkarsh Sinha}, \bibinfo{person}{Jonathan~T. Barron}, \bibinfo{person}{Sofien Bouaziz}, \bibinfo{person}{Dan~B Goldman}, \bibinfo{person}{Steven~M. Seitz}, {and} \bibinfo{person}{Ricardo Martin-Brualla}.} \bibinfo{year}{2021}\natexlab{a}.
\newblock \showarticletitle{Nerfies: Deformable Neural Radiance Fields}. In \bibinfo{booktitle}{\emph{ICCV}}.
\newblock


\bibitem[Park et~al\mbox{.}(2021b)]%
        {park2021hypernerf}
\bibfield{author}{\bibinfo{person}{Keunhong Park}, \bibinfo{person}{Utkarsh Sinha}, \bibinfo{person}{Peter Hedman}, \bibinfo{person}{Jonathan~T Barron}, \bibinfo{person}{Sofien Bouaziz}, \bibinfo{person}{Dan~B Goldman}, \bibinfo{person}{Ricardo Martin-Brualla}, {and} \bibinfo{person}{Steven~M Seitz}.} \bibinfo{year}{2021}\natexlab{b}.
\newblock \showarticletitle{Hypernerf: A higher-dimensional representation for topologically varying neural radiance fields}.
\newblock \bibinfo{journal}{\emph{arXiv preprint arXiv:2106.13228}} (\bibinfo{year}{2021}).
\newblock


\bibitem[Peng et~al\mbox{.}(2023)]%
        {peng2023representing}
\bibfield{author}{\bibinfo{person}{Sida Peng}, \bibinfo{person}{Yunzhi Yan}, \bibinfo{person}{Qing Shuai}, \bibinfo{person}{Hujun Bao}, {and} \bibinfo{person}{Xiaowei Zhou}.} \bibinfo{year}{2023}\natexlab{}.
\newblock \showarticletitle{Representing Volumetric Videos as Dynamic MLP Maps}. In \bibinfo{booktitle}{\emph{CVPR}}.
\newblock


\bibitem[Peng et~al\mbox{.}(2021)]%
        {peng2021neural}
\bibfield{author}{\bibinfo{person}{Sida Peng}, \bibinfo{person}{Yuanqing Zhang}, \bibinfo{person}{Yinghao Xu}, \bibinfo{person}{Qianqian Wang}, \bibinfo{person}{Qing Shuai}, \bibinfo{person}{Hujun Bao}, {and} \bibinfo{person}{Xiaowei Zhou}.} \bibinfo{year}{2021}\natexlab{}.
\newblock \showarticletitle{Neural Body: Implicit Neural Representations with Structured Latent Codes for Novel View Synthesis of Dynamic Humans}. In \bibinfo{booktitle}{\emph{CVPR}}.
\newblock


\bibitem[Penner and Zhang(2017)]%
        {penner2017soft}
\bibfield{author}{\bibinfo{person}{Eric Penner} {and} \bibinfo{person}{Li Zhang}.} \bibinfo{year}{2017}\natexlab{}.
\newblock \showarticletitle{Soft 3D reconstruction for view synthesis}.
\newblock \bibinfo{journal}{\emph{ACM TOG}} (\bibinfo{year}{2017}).
\newblock


\bibitem[Pumarola et~al\mbox{.}(2021)]%
        {pumarola2020d}
\bibfield{author}{\bibinfo{person}{Albert Pumarola}, \bibinfo{person}{Enric Corona}, \bibinfo{person}{Gerard Pons-Moll}, {and} \bibinfo{person}{Francesc Moreno-Noguer}.} \bibinfo{year}{2021}\natexlab{}.
\newblock \showarticletitle{D-NeRF: Neural Radiance Fields for Dynamic Scenes}. In \bibinfo{booktitle}{\emph{CVPR}}.
\newblock


\bibitem[Qi et~al\mbox{.}(2017)]%
        {qi2017pointnet}
\bibfield{author}{\bibinfo{person}{Charles~R Qi}, \bibinfo{person}{Hao Su}, \bibinfo{person}{Kaichun Mo}, {and} \bibinfo{person}{Leonidas~J Guibas}.} \bibinfo{year}{2017}\natexlab{}.
\newblock \showarticletitle{Pointnet: Deep learning on point sets for 3d classification and segmentation}. In \bibinfo{booktitle}{\emph{CVPR}}.
\newblock


\bibitem[Reiser et~al\mbox{.}(2021)]%
        {reiser2021kilonerf}
\bibfield{author}{\bibinfo{person}{Christian Reiser}, \bibinfo{person}{Songyou Peng}, \bibinfo{person}{Yiyi Liao}, {and} \bibinfo{person}{Andreas Geiger}.} \bibinfo{year}{2021}\natexlab{}.
\newblock \showarticletitle{KiloNeRF: Speeding Up Neural Radiance Fields With Thousands of Tiny MLPs}. In \bibinfo{booktitle}{\emph{ICCV}}. \bibinfo{pages}{14335--14345}.
\newblock


\bibitem[Riegler and Koltun(2020)]%
        {Riegler2020FVS}
\bibfield{author}{\bibinfo{person}{Gernot Riegler} {and} \bibinfo{person}{Vladlen Koltun}.} \bibinfo{year}{2020}\natexlab{}.
\newblock \showarticletitle{Free View Synthesis}. In \bibinfo{booktitle}{\emph{ECCV}}.
\newblock


\bibitem[Riegler and Koltun(2021)]%
        {riegler2021stable}
\bibfield{author}{\bibinfo{person}{Gernot Riegler} {and} \bibinfo{person}{Vladlen Koltun}.} \bibinfo{year}{2021}\natexlab{}.
\newblock \showarticletitle{Stable view synthesis}. In \bibinfo{booktitle}{\emph{Proceedings of the IEEE/CVF Conference on Computer Vision and Pattern Recognition}}. \bibinfo{pages}{12216--12225}.
\newblock


\bibitem[Ronneberger et~al\mbox{.}(2015)]%
        {ronneberger2015u}
\bibfield{author}{\bibinfo{person}{Olaf Ronneberger}, \bibinfo{person}{Philipp Fischer}, {and} \bibinfo{person}{Thomas Brox}.} \bibinfo{year}{2015}\natexlab{}.
\newblock \showarticletitle{U-net: Convolutional networks for biomedical image segmentation}. In \bibinfo{booktitle}{\emph{MICCAI}}.
\newblock


\bibitem[Shao et~al\mbox{.}(2022)]%
        {shao2022tensor4d}
\bibfield{author}{\bibinfo{person}{Ruizhi Shao}, \bibinfo{person}{Zerong Zheng}, \bibinfo{person}{Hanzhang Tu}, \bibinfo{person}{Boning Liu}, \bibinfo{person}{Hongwen Zhang}, {and} \bibinfo{person}{Yebin Liu}.} \bibinfo{year}{2022}\natexlab{}.
\newblock \showarticletitle{Tensor4D: Efficient Neural 4D Decomposition for High-fidelity Dynamic Reconstruction and Rendering}.
\newblock \bibinfo{journal}{\emph{arXiv}} (\bibinfo{year}{2022}).
\newblock


\bibitem[Shih et~al\mbox{.}(2020)]%
        {shih20203d}
\bibfield{author}{\bibinfo{person}{Meng-Li Shih}, \bibinfo{person}{Shih-Yang Su}, \bibinfo{person}{Johannes Kopf}, {and} \bibinfo{person}{Jia-Bin Huang}.} \bibinfo{year}{2020}\natexlab{}.
\newblock \showarticletitle{3d photography using context-aware layered depth inpainting}. In \bibinfo{booktitle}{\emph{CVPR}}.
\newblock


\bibitem[Sitzmann et~al\mbox{.}(2021)]%
        {sitzmann2021light}
\bibfield{author}{\bibinfo{person}{Vincent Sitzmann}, \bibinfo{person}{Semon Rezchikov}, \bibinfo{person}{Bill Freeman}, \bibinfo{person}{Josh Tenenbaum}, {and} \bibinfo{person}{Fredo Durand}.} \bibinfo{year}{2021}\natexlab{}.
\newblock \showarticletitle{Light field networks: Neural scene representations with single-evaluation rendering}.
\newblock \bibinfo{journal}{\emph{Advances in Neural Information Processing Systems}}  \bibinfo{volume}{34} (\bibinfo{year}{2021}), \bibinfo{pages}{19313--19325}.
\newblock


\bibitem[Sitzmann et~al\mbox{.}(2019)]%
        {sitzmann2019deepvoxels}
\bibfield{author}{\bibinfo{person}{Vincent Sitzmann}, \bibinfo{person}{Justus Thies}, \bibinfo{person}{Felix Heide}, \bibinfo{person}{Matthias Nie{\ss}ner}, \bibinfo{person}{Gordon Wetzstein}, {and} \bibinfo{person}{Michael Zollh{\"{o}}fer}.} \bibinfo{year}{2019}\natexlab{}.
\newblock \showarticletitle{DeepVoxels: Learning Persistent 3D Feature Embeddings}. In \bibinfo{booktitle}{\emph{CVPR}}.
\newblock
\urldef\tempurl%
\url{https://doi.org/10.1109/CVPR.2019.00254}
\showDOI{\tempurl}


\bibitem[Song et~al\mbox{.}(2023)]%
        {song2023nerfplayer}
\bibfield{author}{\bibinfo{person}{Liangchen Song}, \bibinfo{person}{Anpei Chen}, \bibinfo{person}{Zhong Li}, \bibinfo{person}{Zhang Chen}, \bibinfo{person}{Lele Chen}, \bibinfo{person}{Junsong Yuan}, \bibinfo{person}{Yi Xu}, {and} \bibinfo{person}{Andreas Geiger}.} \bibinfo{year}{2023}\natexlab{}.
\newblock \showarticletitle{Nerfplayer: A streamable dynamic scene representation with decomposed neural radiance fields}.
\newblock \bibinfo{journal}{\emph{IEEE Transactions on Visualization and Computer Graphics}} \bibinfo{volume}{29}, \bibinfo{number}{5} (\bibinfo{year}{2023}), \bibinfo{pages}{2732--2742}.
\newblock


\bibitem[Srinivasan et~al\mbox{.}(2019)]%
        {srinivasan2019pushing}
\bibfield{author}{\bibinfo{person}{Pratul~P Srinivasan}, \bibinfo{person}{Richard Tucker}, \bibinfo{person}{Jonathan~T Barron}, \bibinfo{person}{Ravi Ramamoorthi}, \bibinfo{person}{Ren Ng}, {and} \bibinfo{person}{Noah Snavely}.} \bibinfo{year}{2019}\natexlab{}.
\newblock \showarticletitle{Pushing the boundaries of view extrapolation with multiplane images}. In \bibinfo{booktitle}{\emph{Proceedings of the IEEE/CVF Conference on Computer Vision and Pattern Recognition}}. \bibinfo{pages}{175--184}.
\newblock


\bibitem[Suhail et~al\mbox{.}(2022)]%
        {suhail2022cvpr}
\bibfield{author}{\bibinfo{person}{Mohammed Suhail}, \bibinfo{person}{Carlos Esteves}, \bibinfo{person}{Leonid Sigal}, {and} \bibinfo{person}{Ameesh Makadia}.} \bibinfo{year}{2022}\natexlab{}.
\newblock \showarticletitle{Light Field Neural Rendering}. In \bibinfo{booktitle}{\emph{Proceedings of the IEEE/CVF Conference on Computer Vision and Pattern Recognition (CVPR)}}. \bibinfo{pages}{8269--8279}.
\newblock


\bibitem[Sullivan et~al\mbox{.}(2012)]%
        {sullivan2012overview}
\bibfield{author}{\bibinfo{person}{Gary~J Sullivan}, \bibinfo{person}{Jens-Rainer Ohm}, \bibinfo{person}{Woo-Jin Han}, {and} \bibinfo{person}{Thomas Wiegand}.} \bibinfo{year}{2012}\natexlab{}.
\newblock \showarticletitle{Overview of the high efficiency video coding (HEVC) standard}.
\newblock \bibinfo{journal}{\emph{IEEE Transactions on circuits and systems for video technology}} \bibinfo{volume}{22}, \bibinfo{number}{12} (\bibinfo{year}{2012}), \bibinfo{pages}{1649--1668}.
\newblock


\bibitem[Sun et~al\mbox{.}(2021)]%
        {sun2021direct}
\bibfield{author}{\bibinfo{person}{Cheng Sun}, \bibinfo{person}{Min Sun}, {and} \bibinfo{person}{Hwann-Tzong Chen}.} \bibinfo{year}{2021}\natexlab{}.
\newblock \showarticletitle{Direct Voxel Grid Optimization: Super-fast Convergence for Radiance Fields Reconstruction}.
\newblock \bibinfo{journal}{\emph{arXiv preprint arXiv:2111.11215}} (\bibinfo{year}{2021}).
\newblock


\bibitem[Sun et~al\mbox{.}(2020)]%
        {sun2020light}
\bibfield{author}{\bibinfo{person}{Tiancheng Sun}, \bibinfo{person}{Zexiang Xu}, \bibinfo{person}{Xiuming Zhang}, \bibinfo{person}{Sean Fanello}, \bibinfo{person}{Christoph Rhemann}, \bibinfo{person}{Paul Debevec}, \bibinfo{person}{Yun-Ta Tsai}, \bibinfo{person}{Jonathan~T Barron}, {and} \bibinfo{person}{Ravi Ramamoorthi}.} \bibinfo{year}{2020}\natexlab{}.
\newblock \showarticletitle{Light stage super-resolution: continuous high-frequency relighting}.
\newblock \bibinfo{journal}{\emph{ACM Transactions on Graphics (TOG)}} \bibinfo{volume}{39}, \bibinfo{number}{6} (\bibinfo{year}{2020}), \bibinfo{pages}{1--12}.
\newblock


\bibitem[Szeliski and Golland(1998)]%
        {szeliski1998stereo}
\bibfield{author}{\bibinfo{person}{Richard Szeliski} {and} \bibinfo{person}{Polina Golland}.} \bibinfo{year}{1998}\natexlab{}.
\newblock \showarticletitle{Stereo matching with transparency and matting}. In \bibinfo{booktitle}{\emph{Sixth International Conference on Computer Vision (IEEE Cat. No. 98CH36271)}}. IEEE, \bibinfo{pages}{517--524}.
\newblock


\bibitem[Tancik et~al\mbox{.}(2022)]%
        {tancik2022block}
\bibfield{author}{\bibinfo{person}{Matthew Tancik}, \bibinfo{person}{Vincent Casser}, \bibinfo{person}{Xinchen Yan}, \bibinfo{person}{Sabeek Pradhan}, \bibinfo{person}{Ben Mildenhall}, \bibinfo{person}{Pratul~P Srinivasan}, \bibinfo{person}{Jonathan~T Barron}, {and} \bibinfo{person}{Henrik Kretzschmar}.} \bibinfo{year}{2022}\natexlab{}.
\newblock \showarticletitle{Block-nerf: Scalable large scene neural view synthesis}. In \bibinfo{booktitle}{\emph{CVPR}}.
\newblock


\bibitem[Tucker and Snavely(2020)]%
        {tucker2020single}
\bibfield{author}{\bibinfo{person}{Richard Tucker} {and} \bibinfo{person}{Noah Snavely}.} \bibinfo{year}{2020}\natexlab{}.
\newblock \showarticletitle{Single-view view synthesis with multiplane images}. In \bibinfo{booktitle}{\emph{Proceedings of the IEEE/CVF Conference on Computer Vision and Pattern Recognition}}. \bibinfo{pages}{551--560}.
\newblock


\bibitem[Turki et~al\mbox{.}(2022)]%
        {turki2022mega}
\bibfield{author}{\bibinfo{person}{Haithem Turki}, \bibinfo{person}{Deva Ramanan}, {and} \bibinfo{person}{Mahadev Satyanarayanan}.} \bibinfo{year}{2022}\natexlab{}.
\newblock \showarticletitle{Mega-NeRF: Scalable Construction of Large-Scale NeRFs for Virtual Fly-Throughs}. In \bibinfo{booktitle}{\emph{CVPR}}.
\newblock


\bibitem[Wan et~al\mbox{.}(2023)]%
        {wan2023learning}
\bibfield{author}{\bibinfo{person}{Ziyu Wan}, \bibinfo{person}{Christian Richardt}, \bibinfo{person}{Alja{\v{z}} Bo{\v{z}}i{\v{c}}}, \bibinfo{person}{Chao Li}, \bibinfo{person}{Vijay Rengarajan}, \bibinfo{person}{Seonghyeon Nam}, \bibinfo{person}{Xiaoyu Xiang}, \bibinfo{person}{Tuotuo Li}, \bibinfo{person}{Bo Zhu}, \bibinfo{person}{Rakesh Ranjan}, {et~al\mbox{.}}} \bibinfo{year}{2023}\natexlab{}.
\newblock \showarticletitle{Learning Neural Duplex Radiance Fields for Real-Time View Synthesis}.
\newblock \bibinfo{journal}{\emph{arXiv preprint arXiv:2304.10537}} (\bibinfo{year}{2023}).
\newblock


\bibitem[Wang et~al\mbox{.}(2022)]%
        {wang2022fourier}
\bibfield{author}{\bibinfo{person}{Liao Wang}, \bibinfo{person}{Jiakai Zhang}, \bibinfo{person}{Xinhang Liu}, \bibinfo{person}{Fuqiang Zhao}, \bibinfo{person}{Yanshun Zhang}, \bibinfo{person}{Yingliang Zhang}, \bibinfo{person}{Minye Wu}, \bibinfo{person}{Lan Xu}, {and} \bibinfo{person}{Jingyi Yu}.} \bibinfo{year}{2022}\natexlab{}.
\newblock \showarticletitle{Fourier PlenOctrees for Dynamic Radiance Field Rendering in Real-time}.
\newblock \bibinfo{journal}{\emph{CVPR}} (\bibinfo{year}{2022}).
\newblock


\bibitem[Wang et~al\mbox{.}(2021)]%
        {wang2021ibrnet}
\bibfield{author}{\bibinfo{person}{Qianqian Wang}, \bibinfo{person}{Zhicheng Wang}, \bibinfo{person}{Kyle Genova}, \bibinfo{person}{Pratul Srinivasan}, \bibinfo{person}{Howard Zhou}, \bibinfo{person}{Jonathan~T. Barron}, \bibinfo{person}{Ricardo Martin-Brualla}, \bibinfo{person}{Noah Snavely}, {and} \bibinfo{person}{Thomas Funkhouser}.} \bibinfo{year}{2021}\natexlab{}.
\newblock \showarticletitle{IBRNet: Learning Multi-View Image-Based Rendering}. In \bibinfo{booktitle}{\emph{CVPR}}.
\newblock


\bibitem[Wang et~al\mbox{.}(2023)]%
        {wang2023inv}
\bibfield{author}{\bibinfo{person}{Shengze Wang}, \bibinfo{person}{Alexey Supikov}, \bibinfo{person}{Joshua Ratcliff}, \bibinfo{person}{Henry Fuchs}, {and} \bibinfo{person}{Ronald Azuma}.} \bibinfo{year}{2023}\natexlab{}.
\newblock \showarticletitle{INV: Towards Streaming Incremental Neural Videos}.
\newblock \bibinfo{journal}{\emph{arXiv preprint arXiv:2302.01532}} (\bibinfo{year}{2023}).
\newblock


\bibitem[Wizadwongsa et~al\mbox{.}(2021)]%
        {wizadwongsa2021nex}
\bibfield{author}{\bibinfo{person}{Suttisak Wizadwongsa}, \bibinfo{person}{Pakkapon Phongthawee}, \bibinfo{person}{Jiraphon Yenphraphai}, {and} \bibinfo{person}{Supasorn Suwajanakorn}.} \bibinfo{year}{2021}\natexlab{}.
\newblock \showarticletitle{Nex: Real-time view synthesis with neural basis expansion}. In \bibinfo{booktitle}{\emph{CVPR}}.
\newblock


\bibitem[Wu et~al\mbox{.}(2020)]%
        {wu2020multi}
\bibfield{author}{\bibinfo{person}{Minye Wu}, \bibinfo{person}{Yuehao Wang}, \bibinfo{person}{Qiang Hu}, {and} \bibinfo{person}{Jingyi Yu}.} \bibinfo{year}{2020}\natexlab{}.
\newblock \showarticletitle{Multi-View Neural Human Rendering}. In \bibinfo{booktitle}{\emph{CVPR}}.
\newblock


\bibitem[Xian et~al\mbox{.}(2021)]%
        {xian2021space}
\bibfield{author}{\bibinfo{person}{Wenqi Xian}, \bibinfo{person}{Jia-Bin Huang}, \bibinfo{person}{Johannes Kopf}, {and} \bibinfo{person}{Changil Kim}.} \bibinfo{year}{2021}\natexlab{}.
\newblock \showarticletitle{Space-time neural irradiance fields for free-viewpoint video}. In \bibinfo{booktitle}{\emph{Proceedings of the IEEE/CVF Conference on Computer Vision and Pattern Recognition}}. \bibinfo{pages}{9421--9431}.
\newblock


\bibitem[You and Hou(2023)]%
        {you2023decoupling}
\bibfield{author}{\bibinfo{person}{Meng You} {and} \bibinfo{person}{Junhui Hou}.} \bibinfo{year}{2023}\natexlab{}.
\newblock \showarticletitle{Decoupling Dynamic Monocular Videos for Dynamic View Synthesis}.
\newblock \bibinfo{journal}{\emph{arXiv preprint arXiv:2304.01716}} (\bibinfo{year}{2023}).
\newblock


\bibitem[Yu et~al\mbox{.}(2022)]%
        {yu2021plenoxels}
\bibfield{author}{\bibinfo{person}{Alex Yu}, \bibinfo{person}{Sara Fridovich-Keil}, \bibinfo{person}{Matthew Tancik}, \bibinfo{person}{Qinhong Chen}, \bibinfo{person}{Benjamin Recht}, {and} \bibinfo{person}{Angjoo Kanazawa}.} \bibinfo{year}{2022}\natexlab{}.
\newblock \showarticletitle{Plenoxels: Radiance Fields without Neural Networks}.
\newblock \bibinfo{journal}{\emph{CVPR}} (\bibinfo{year}{2022}).
\newblock


\bibitem[Yu et~al\mbox{.}(2021a)]%
        {yu2021plenoctrees}
\bibfield{author}{\bibinfo{person}{Alex Yu}, \bibinfo{person}{Ruilong Li}, \bibinfo{person}{Matthew Tancik}, \bibinfo{person}{Hao Li}, \bibinfo{person}{Ren Ng}, {and} \bibinfo{person}{Angjoo Kanazawa}.} \bibinfo{year}{2021}\natexlab{a}.
\newblock \showarticletitle{PlenOctrees for Real-Time Rendering of Neural Radiance Fields}. In \bibinfo{booktitle}{\emph{ICCV}}.
\newblock


\bibitem[Yu et~al\mbox{.}(2021b)]%
        {yu2020pixelnerf}
\bibfield{author}{\bibinfo{person}{Alex Yu}, \bibinfo{person}{Vickie Ye}, \bibinfo{person}{Matthew Tancik}, {and} \bibinfo{person}{Angjoo Kanazawa}.} \bibinfo{year}{2021}\natexlab{b}.
\newblock \showarticletitle{{pixelNeRF}: Neural Radiance Fields from One or Few Images}. In \bibinfo{booktitle}{\emph{CVPR}}.
\newblock


\bibitem[Yu et~al\mbox{.}(2018)]%
        {yu2018doublefusion}
\bibfield{author}{\bibinfo{person}{Tao Yu}, \bibinfo{person}{Zerong Zheng}, \bibinfo{person}{Kaiwen Guo}, \bibinfo{person}{Jianhui Zhao}, \bibinfo{person}{Qionghai Dai}, \bibinfo{person}{Hao Li}, \bibinfo{person}{Gerard Pons-Moll}, {and} \bibinfo{person}{Yebin Liu}.} \bibinfo{year}{2018}\natexlab{}.
\newblock \showarticletitle{Doublefusion: Real-time capture of human performances with inner body shapes from a single depth sensor}. In \bibinfo{booktitle}{\emph{CVPR}}.
\newblock


\bibitem[Zhang et~al\mbox{.}(2021)]%
        {zhang2021editable}
\bibfield{author}{\bibinfo{person}{Jiakai Zhang}, \bibinfo{person}{Xinhang Liu}, \bibinfo{person}{Xinyi Ye}, \bibinfo{person}{Fuqiang Zhao}, \bibinfo{person}{Yanshun Zhang}, \bibinfo{person}{Minye Wu}, \bibinfo{person}{Yingliang Zhang}, \bibinfo{person}{Lan Xu}, {and} \bibinfo{person}{Jingyi Yu}.} \bibinfo{year}{2021}\natexlab{}.
\newblock \showarticletitle{Editable free-viewpoint video using a layered neural representation}.
\newblock \bibinfo{journal}{\emph{ACM Transactions on Graphics (TOG)}} \bibinfo{volume}{40}, \bibinfo{number}{4} (\bibinfo{year}{2021}), \bibinfo{pages}{1--18}.
\newblock


\bibitem[Zhou et~al\mbox{.}(2018)]%
        {zhou2018stereo}
\bibfield{author}{\bibinfo{person}{Tinghui Zhou}, \bibinfo{person}{Richard Tucker}, \bibinfo{person}{John Flynn}, \bibinfo{person}{Graham Fyffe}, {and} \bibinfo{person}{Noah Snavely}.} \bibinfo{year}{2018}\natexlab{}.
\newblock \showarticletitle{Stereo magnification: Learning view synthesis using multiplane images}. In \bibinfo{booktitle}{\emph{SIGGRAPH}}.
\newblock


\bibitem[Zitnick et~al\mbox{.}(2004)]%
        {zitnick2004high}
\bibfield{author}{\bibinfo{person}{C~Lawrence Zitnick}, \bibinfo{person}{Sing~Bing Kang}, \bibinfo{person}{Matthew Uyttendaele}, \bibinfo{person}{Simon Winder}, {and} \bibinfo{person}{Richard Szeliski}.} \bibinfo{year}{2004}\natexlab{}.
\newblock \showarticletitle{High-quality video view interpolation using a layered representation}.
\newblock \bibinfo{journal}{\emph{ACM TOG}} (\bibinfo{year}{2004}).
\newblock


\end{thebibliography}
\newpage
\appendix

\setcounter{section}{0}
\section*{Supplementary material}
In the supplementary material, we provide implementation details, more visualization results and per-scene breakdown. Please refer to our video for more comparison results and visualization results.

\section{Implementation Details}
\subsection{Loss Formulation}
We optimize our scene representation using the mean squared error (MSE) loss with color images. 
\begin{equation}
    \mathcal{L}_\text{mse} = \frac{1}{\numPixel} \sum_{i=1}^{\numPixel} (\colorpixel_i - \hat{\colorpixel_i})^2,
\end{equation}
where $\numPixel$ represents the number of sampled pixels for each training iteration, and $\colorpixel_i$ and $\hat{\colorpixel_i}$ are the ground truth and predicted colors of the $i$-th pixel, respectively.
In addition to the color loss, we follow \cite{yu2021plenoxels,fridovich2023k} and use the total variation (TV) loss to regularize the explicit feature planes $\{\plane_i | i \in \{xy, xz, yz, xt, yt, zt\}\}$. 
We detail our loss formulation in the supplementary material.
The TV loss can be formulated as:
\begin{equation}
    \mathcal{L}_\text{tv} (\plane) = \sum_{j=1}^{\planesizeh-1} \sum_{k=1}^{\planesizew-1} \left(\plane (j+1, k) - \plane (j, k))^2 + (\plane (j, k+1) - \plane (j, k))^2 \right), 
\end{equation}
where $\planesizeh$ and $\planesizew$ denote the height and width of the explicit feature plane $\plane$, respectively.
We sum the TV losses for the six feature planes as the final TV loss.
The TV loss is assigned a weight of 0.001, while the MSE loss has a weight of 1 in the final loss calculation.

\subsection{Evaluation Details}
\begin{figure}[t]
    \centering
    \includegraphics[width=1\columnwidth]{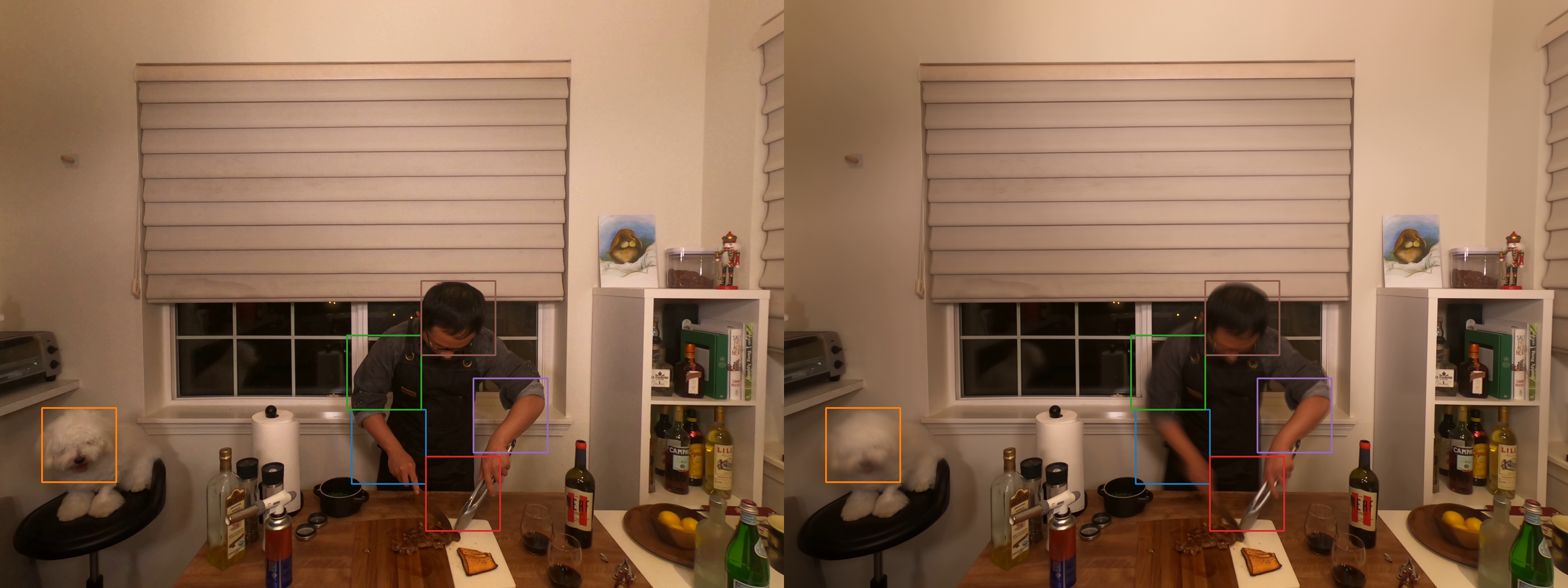}
    \caption{\textbf{Evaluation details on the DyNeRF dataset.} The left image is the first frame (test frame) of a one-second video clip, and the right image is the average frame of this second. We identify the 6 patches with the largest differences between the test frame and the average frame. During the quantitative evaluation, we only assess these patches.}
    \label{fig:dynerf_vis}
\end{figure}

We include the evaluation details for the DyNeRF dataset \cite{li2021neural} in Fig. \ref{fig:dynerf_vis}.
For a 10-second clip, we evaluate 10 frames and Fig. \ref{fig:dynerf_vis} presents one of the frames.

\subsection{Other Details}

Our feature planes have mulptle spatial $(xy,yz,xz)$ resolutions ($512\times512$, $256\times256$, $128\times128$, $64\times64$) and one temporal $(xt, yt, zt)$ resolution ($spatial\_resolution \times \frac{frame\_number}{2}$).
The number of channels for each feature plane is 16.
These feature planes 
We use concatenation to aggregate multi-resolution features.
The CNN feature network's architecture is the same as the architecture used in ENeRF \cite{lin2022enerf}.
The hidden size of $\mlpApp$ and $\mlpGeo$ is 64, and the number of hidden layers is 1. We use ReLU as the activation function.
Our appearance-related networks can also be pre-trained with generalizable radiance field methods\cite{lin2022enerf,wang2021ibrnet}. 
Specifically, we slightly modify the appearance rendering head of ENeRF \cite{lin2022enerf} to align it with our method and retrain it on the DTU dataset. 
To train our method on a new scene, we will load this pre-trained appearance network.
We experimentally found that this helps our method get slight better results as shown in the ablations in the main paper.
We optimize our model using Adam \cite{kingma2014adam} optimizer with an initial learning rate of 0.01 for explicit plane features and 0.001 for MLPs and the feature network.
The learning rate is decayed with a cosine learning rate decay strategy and will be decayed into zero when training is finished. (100k iterations on the ZJU-MoCap and NHR datasets, 140k on the DNA-Rendering datasets, 15K on the DyNeRF dataset.)
For each training iteration, we randomly select 4 images and sample 1024 pixels from each image as a training batch.
Following \cite{barron2021mip}, we use an additional proposal $\mlpGeo$ with feature planes of a single spatial resolution (128) to propose points.
For each ray, we uniformly sample 64 points to infer the proposal network to get 32 proposal points. 
We implement our efficient rendering strategy using NerfAcc \cite{li2023nerfacc}.

\begin{figure}
    \centering
    \includegraphics[width=\columnwidth]{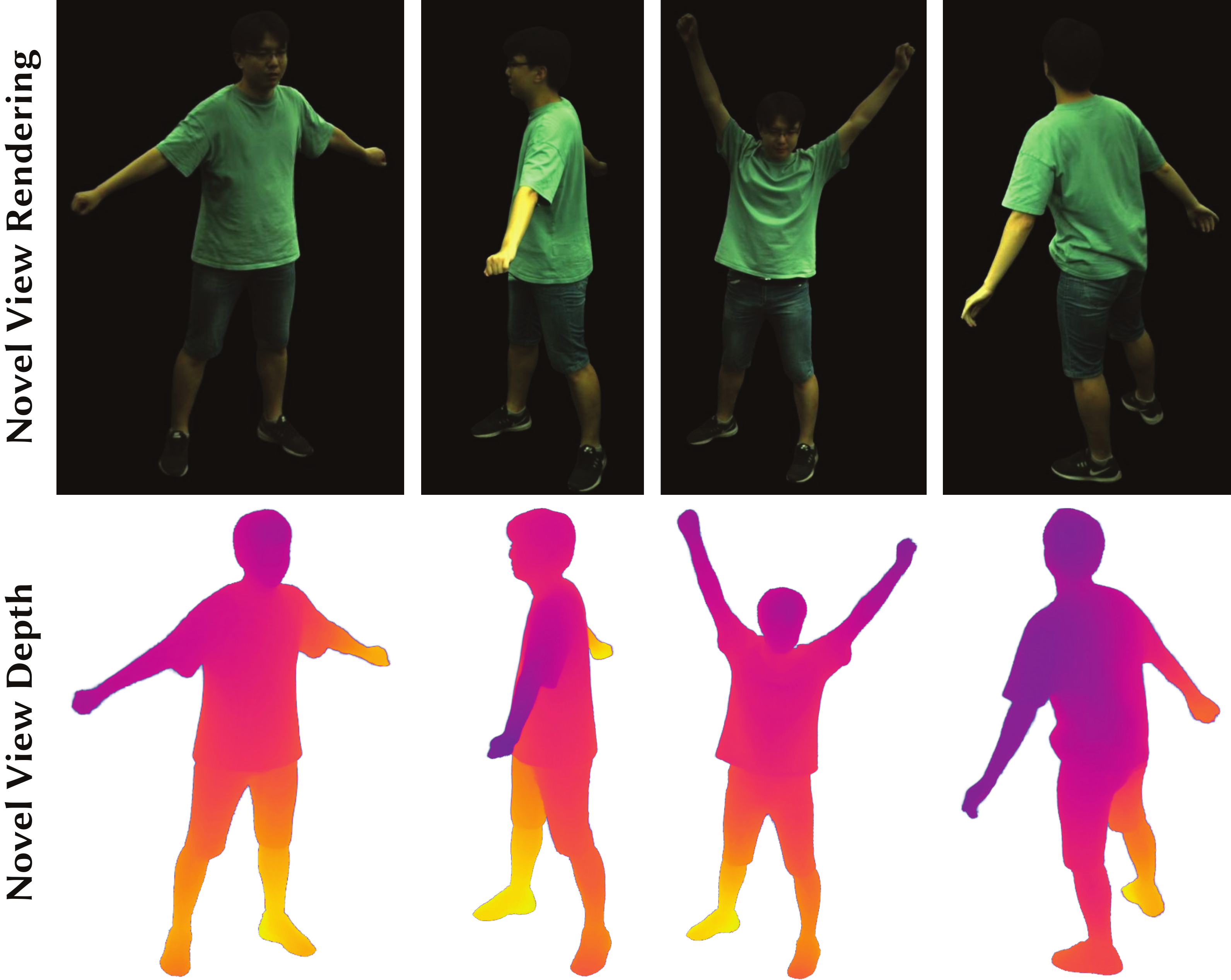}
    \caption{\textbf{Novel view synthesis results on the ZJU-MoCap dataset.} These images are taken from a video generated by rendering a trajectory from novel perspectives. Please watch the complete video sequence in the supplementary video. The high-quality depth results indicate that our method can achieve inter-view rendering consistency.}
    \label{fig:depth_zjumocap}
\end{figure}

\begin{figure*}
    \centering
    \includegraphics[width=0.95\linewidth]{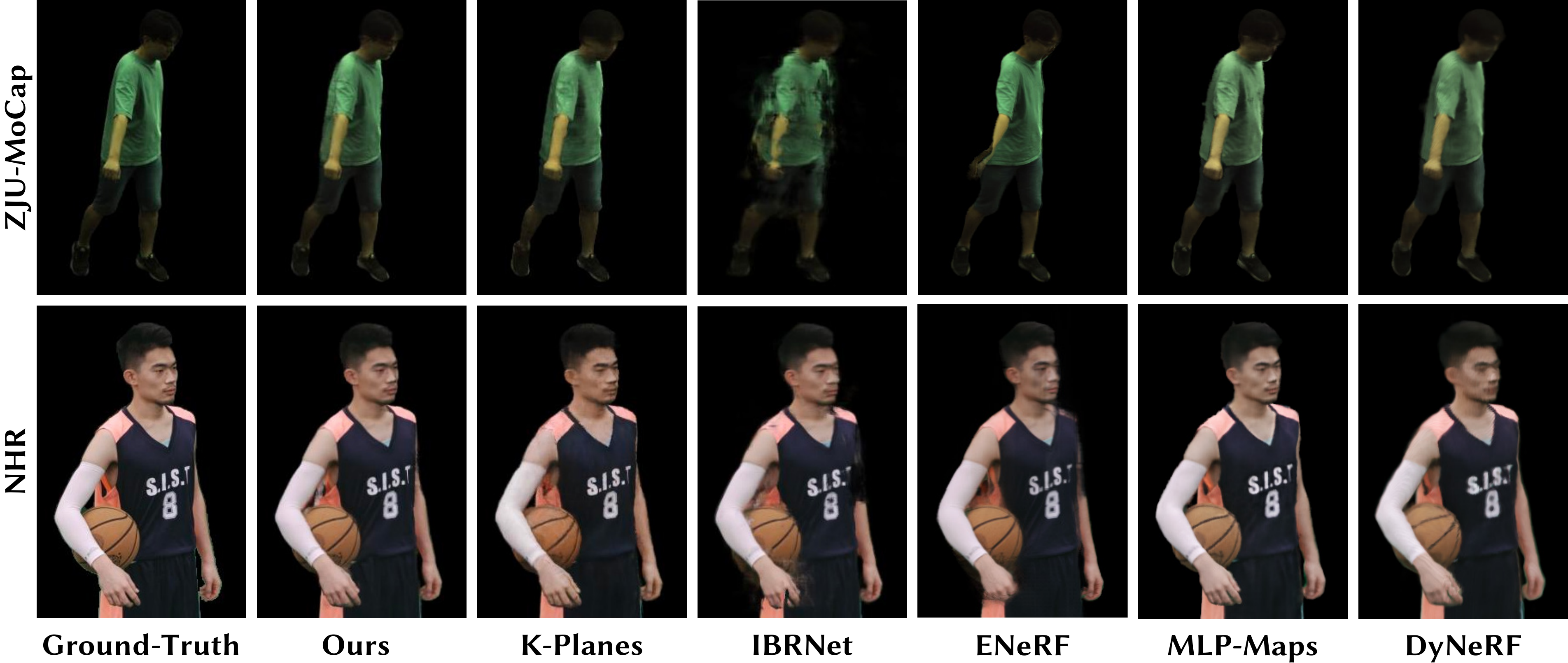}
    \vspace{-2mm}
    \caption{\textbf{Qualitative comparison of image synthesis results on the ZJU-MoCap and NHR datasets.} 
    }
    \label{fig:comp_zjuvis}
    \vspace{-2mm}
\end{figure*}

\section{Additional Results}

We include qualitative comparison results on the ZJU-MoCap and NHR datasets in Fig.\ref{fig:comp_zjuvis}.
We provide additional results on the ZJU-MoCap dataset in Figure \ref{fig:depth_zjumocap}. 
The high-quality depth results we achieved are the reason why our method is able to provide consistent rendering.

\section{Per-scene breakdown}
Tables \ref{tab:renbody}, \ref{tab:zjumocap_nhr} present the per-scene comparisons.
These results are consistent with the averaged results in the paper.


\begin{table*}
	\centering
	\caption{\textbf{Quantitative comparison of view synthesis results on the DNA-Rendering dataset.}}
	\begin{tabular}{l|cccc|cccc|cccc}
		\Xhline{3\arrayrulewidth}
        & 0008\_03 & 0013\_01 & 0013\_03 & 0013\_09 & 0008\_03 & 0013\_01 & 0013\_03 & 0013\_09 & 0008\_03 & 0013\_01 & 0013\_03 & 0013\_09 \\ \hline

		& \multicolumn{4}{c|}{PSNR$\uparrow$} & \multicolumn{4}{c|}{SSIM$\uparrow$} & \multicolumn{4}{c}{LPIPS$\downarrow$} \\ \hline
		K-Planes$_{2h}$ & 26.94 & 26.00 & 25.91 & 26.53 & 0.930 & 0.943 & 0.944 & 0.957 & 0.185 & 0.118 & 0.114 & 0.117 \\
        K-Planes$_{8h}$ & \cellsecond 27.99 & 26.80 & 27.87 & 27.15 & 0.940 & 0.950 & 0.958 & 0.960 & 0.165 & 0.107 & 0.091 & 0.108 \\ \hline
		IBRNet & 25.76 & 26.08 & 25.15 & 24.28 & 0.945 & 0.960 & 0.952 & 0.959 & 0.150 & 0.087 & 0.091 & 0.097 \\
        IBRNet$_{3.5h}$ & 27.66 & 28.84 & 27.53 & 27.35 & 0.955 & 0.973 & 0.965 & \cellsecond 0.973 & 0.127 & 0.060 & 0.067 & 0.070 \\ \hline
        ENeRF & 26.79 & 27.92 & 26.84 & 25.88 & 0.955 & 0.973 & 0.965 & 0.969 & \cellsecond 0.109 & 0.055 & \cellsecond 0.061 & 0.069 \\
        ENeRF$_{3.5h}$ & 27.74 & \cellsecond 28.99 & \cellsecond 28.08 & \cellsecond 27.84 & \cellsecond 0.958 & \cellsecond 0.976 & \cellsecond 0.970 & 0.970 & 0.113 & \cellsecond 0.048 & \cellfirst 0.052 & \cellsecond 0.052 \\ \hline
        Ours$_{0.51h}$ & 27.70 & 27.62 & 26.46 & 26.80 & 0.952 & 0.966 & 0.960 & 0.968 & 0.121 & 0.067 & 0.069 & 0.077 \\
        Ours$_{3.33h}$ & \cellfirst 28.90 & \cellfirst 29.53 & \cellfirst 28.75 & \cellfirst 28.78 & \cellfirst 0.962 & \cellfirst 0.977 & \cellfirst 0.972 & \cellfirst 0.979 & \cellfirst 0.096 & \cellfirst 0.047 & \cellfirst 0.052 & \cellfirst 0.051 \\
        \Xhline{3\arrayrulewidth}

\end{tabular}

\label{tab:renbody}
\end{table*}


\begin{table*}
	\centering
	\caption{\textbf{Quantitative comparison of view synthesis results on ZJU-MoCap and NHR datasets.}}
	\begin{tabular}{l|ccccccccc|cccc}
		\Xhline{3\arrayrulewidth}
        & \multicolumn{9}{c|}{ZJU-MoCap} & \multicolumn{4}{c}{NHR} \\
        \hline
        & 313 & 315 & 377 & 386 & 387 & 390 & 392 & 393 & 394 & basketball & sport\_1 & sport\_2 & sport\_3 \\ \hline

		& \multicolumn{9}{c|}{PSNR$\uparrow$} & \multicolumn{4}{c}{PSNR$\uparrow$} \\ \hline
        DyNeRF$_{>24h}$ & 31.50 & 30.29 & 28.92 & \cellsecond 30.88 & \cellsecond 27.90 & 30.14 & 30.09 & 29.28 & 29.88 & 27.97 & 31.76 & 32.43 & 31.33 \\ \hline
		MLP-Maps$_{>24h}$ & \cellsecond 32.15 & 29.94 & \cellsecond 29.40 & 31.05 & 27.89 & 30.10 & \cellsecond 31.06 & \cellsecond 29.78 & 30.15 & 29.11 & 32.92 & 33.19 & 33.59 \\ \hline
        K-Planes$_{2h}$ & 30.85 & 29.23 & 28.69 & 30.67 & 27.43 & 29.84 & 30.12 & 28.85 & 29.82 & 28.26 & 32.23 & 30.70 & 30.89 \\
        K-Planes$_{8h}$ & 32.11 & \cellsecond 30.55 & \cellsecond 29.40 & 30.82 & 27.75 & 30.06 & 31.00 & 29.61 & 30.17 & 30.01 & 34.52 & 33.96 & 33.24 \\ \hline
		IBRNet & 29.08 & 25.13 & 27.47 & 29.97 & 26.27 & 28.59 & 28.90 & 27.86 & 28.18 & 27.01 & 28.91 & 29.94 & 28.68 \\
        IBRNet$_{2.5h}$ & 30.89 & 28.12 & 27.47 & 30.66 & 27.43 & 29.84 & 30.12 & 28.85 & 29.82 & \cellsecond 30.47 & \cellsecond 34.65 & \cellsecond 35.02 & \cellfirst 34.01 \\ \hline
        ENeRF & 30.31 & 28.13 & 28.73 & 30.34 & 27.24 & 29.32 & 29.86 & 28.84 & 29.18 & 25.98 & 25.87 & 27.40 & 26.30 \\
        ENeRF$_{2.5h}$ & 30.69 & 28.79 & 28.51 & 30.10 & 27.32 & 29.09 & 30.03 & 29.07 & 29.29 & 28.22 & 30.68 & 32.08 & 31.26 \\ \hline
        Ours$_{0.49h}$ & 31.62 & 30.09 & 29.24 & \cellfirst 30.89 & 27.89 & \cellsecond 30.16 & 30.84 & 29.59 & \cellsecond 30.27 & 29.51 & 33.67 & 34.21 & 32.21 \\
        Ours$_{2.29h}$ & \cellfirst 32.87 & \cellfirst 30.75 & \cellfirst 29.61 & 30.82 & \cellfirst 28.16 & \cellfirst 30.31 & \cellfirst 31.38 & \cellfirst 30.00 & \cellfirst 30.52 & \cellfirst 30.57 & \cellfirst 34.87 & \cellfirst 35.61 & \cellsecond 33.82 \\
        \Xhline{3\arrayrulewidth}

		& \multicolumn{9}{c|}{SSIM$\uparrow$} & \multicolumn{4}{c}{SSIM$\uparrow$} \\ \hline
        DyNeRF$_{>24h}$ & 0.970 & 0.976 & 0.960 & \cellfirst 0.960 & 0.953 & 0.959 & 0.953 & 0.952 & 0.952 & 0.929 & 0.954 & 0.945 & 0.944 \\ \hline
		MLP-Maps$_{>24h}$ & 0.976 & 0.977 & 0.963 & \cellfirst 0.960 & 0.953 & 0.959 & \cellsecond 0.962 & \cellsecond 0.958 & 0.957 & 0.943 & 0.959 & 0.954 & 0.956 \\ \hline
        K-Planes$_{2h}$ & 0.968 & 0.975 & 0.960 & 0.955 & 0.947 & 0.955 & 0.954 & 0.948 & 0.950 & 0.933 & 0.958 & 0.944 & 0.940 \\
        K-Planes$_{8h}$ & 0.975 & \cellsecond 0.981 & \cellsecond 0.965 & 0.957 & 0.952 & 0.957 & 0.960 & 0.955 & 0.955 & 0.949 & 0.968 & 0.960 & 0.956 \\ \hline
		IBRNet & 0.941 & 0.930 & 0.939 & 0.942 & 0.931 & 0.936 & 0.936 & 0.931 & 0.926 & 0.926 & 0.945 & 0.933 & 0.935 \\
        IBRNet$_{2.5h}$ & 0.968 & 0.968 & 0.960 & 0.954 & 0.948 & 0.952 & 0.958 & 0.951 & 0.946 & \cellsecond 0.961 & \cellsecond 0.969 & \cellsecond 0.964 & \cellsecond 0.965 \\ \hline
        ENeRF & 0.968 & 0.969 & 0.964 & 0.958 & 0.955 & 0.956 & 0.959 & 0.956 & 0.954 & 0.927 & 0.943 & 0.923 & 0.930 \\
        ENeRF$_{2.5h}$ & 0.970 & 0.971 & 0.960 & 0.954 & 0.953 & 0.953 & 0.959 & 0.956 & 0.955 & 0.938 & 0.956 & 0.947 & 0.949 \\ \hline
        Ours$_{0.49h}$ & \cellsecond 0.977 & \cellsecond 0.981 & \cellsecond 0.965 & \cellfirst 0.960 & \cellsecond 0.956 & \cellsecond 0.960 & \cellsecond 0.962 & 0.957 & \cellsecond 0.958 & 0.958 & \cellsecond 0.969 & 0.963 & 0.960 \\
        Ours$_{2.29h}$ & \cellfirst 0.982 & \cellfirst 0.983 & \cellfirst 0.968 & \cellsecond 0.959 & \cellfirst 0.958 & \cellfirst 0.961 & \cellfirst 0.965 & \cellfirst 0.960 & \cellfirst 0.960 & \cellfirst 0.966 & \cellfirst 0.976 & \cellfirst 0.971 & \cellfirst 0.968 \\
        \Xhline{3\arrayrulewidth}

		& \multicolumn{9}{c|}{LPIPS$\downarrow$} & \multicolumn{4}{c}{LPIPS$\downarrow$} \\ \hline
        DyNeRF$_{>24h}$ & 0.070 & 0.061 & 0.083 & 0.082 & 0.094 & 0.083 & 0.113 & 0.102 & 0.099 & 0.142 & 0.095 & 0.119 & 0.114 \\ \hline
		MLP-Maps$_{>24h}$ & 0.049 & 0.047 & 0.069 & 0.072 & 0.081 & 0.068 & 0.074 & 0.080 & 0.072 & 0.094 & 0.067 & 0.084 & 0.076 \\ \hline
        K-Planes$_{2h}$ & 0.085 & 0.070 & 0.106 & 0.125 & 0.117 & 0.100 & 0.124 & 0.118 & 0.117 & 0.164 & 0.103 & 0.140 & 0.135 \\
        K-Planes$_{8h}$ & 0.056 & 0.045 & 0.081 & 0.108 & 0.094 & 0.086 & 0.090 & 0.090 & 0.091 & 0.133 & 0.076 & 0.097 & 0.099 \\ \hline
		IBRNet & 0.116 & 0.124 & 0.121 & 0.132 & 0.126 & 0.129 & 0.132 & 0.126 & 0.129 & 0.123 & 0.094 & 0.123 & 0.114 \\
        IBRNet$_{2.5h}$ & 0.066 & 0.070 & 0.078 & 0.094 & 0.093 & 0.095 & 0.089 & 0.090 & 0.083 & \cellsecond 0.079 & 0.068 & 0.087 & 0.077 \\ \hline
        ENeRF & 0.042 & 0.043 & \cellfirst 0.046 & \cellsecond 0.054 & \cellfirst 0.053 & \cellfirst 0.056 & \cellsecond 0.056 & \cellsecond 0.056 & \cellsecond 0.053 & 0.095 & 0.072 & 0.098 & 0.088 \\
        ENeRF$_{2.5h}$ & \cellsecond 0.036 & 0.039 & \cellfirst 0.046 & \cellfirst 0.052 & \cellfirst 0.053 & \cellfirst 0.056 & \cellfirst 0.051 & \cellfirst 0.055 & \cellfirst 0.049 & 0.088 & 0.061 & \cellsecond 0.077 & \cellsecond 0.075 \\ \hline
        Ours$_{0.49h}$ & 0.040 & \cellsecond 0.034 & \cellsecond 0.063 & 0.068 & 0.068 & \cellsecond 0.064 & 0.071 & 0.073 & 0.068 & 0.081 & \cellsecond 0.059 & 0.079 & \cellsecond 0.075 \\
        Ours$_{2.29h}$ & \cellfirst 0.027 & \cellfirst 0.027 & \cellfirst 0.046 & 0.063 & \cellsecond 0.055 & \cellfirst 0.056 & 0.055 & 0.058 & 0.055 & \cellfirst 0.061 & \cellfirst 0.043 & \cellfirst 0.060 & \cellfirst 0.058 \\
        \Xhline{3\arrayrulewidth}

\end{tabular}

\label{tab:zjumocap_nhr}
\end{table*}

\end{document}